\documentclass[twoside,11pt]{article}

\usepackage{jmlr2e}

\usepackage{amsfonts}
\usepackage{amsbsy}
\usepackage{amsmath}
\usepackage{multirow}
\usepackage{float}
\usepackage[all]{xy}
\usepackage{lscape}

\def\sgn{{\ensuremath {\rm sgn}}}
\newcommand{\IG}{\includegraphics}

\jmlrheading{9}{2008}{49-66}{7/07; Revised 11/07}{1/08}{Gustavo Camps-Valls, Juan
Guti\'errez, Gabriel G\'omez-P\'erez and Jes\'us Malo}

\ShortHeadings{On the Suitable Domain for SVM Training in Image Coding}{Camps-Valls, Guti\'errez, G\'omez-P\'erez and Malo}
\firstpageno{49}

\begin{document}

\title{On the Suitable Domain for SVM Training in Image Coding}

\author{\name Gustavo Camps-Valls \email gustavo.camps@uv.es \\
       \addr Dept. Enginyeria Electr\`onica\\
       Universitat de Val\`encia \\
       46100 Burjassot, Val\`encia, Spain
       \AND
       \name Juan Guti\'errez \email juan.gutierrez@uv.es \\
       \addr Dept. Inform\`atica \\
       Universitat de Val\`encia \\
       46100 Burjassot, Val\`encia, Spain
       \AND
       \name Gabriel G\'omez-P\'erez \email gabriel.gomez@analog.com \\
       \addr Audio and Video Group \\
       Analog Devices Inc. \\
       Limerick, Ireland
       \AND
       \name Jes\'us Malo \email jesus.malo@uv.es \\
       \addr Dept. d'\`Optica \\
       Universitat de Val\`encia \\
       46100 Burjassot, Val\`encia, Spain}

\editor{Donald Geman}

\maketitle

\begin{abstract}
Conventional SVM-based image coding methods are founded on
independently restricting the distortion in every image
coefficient at some particular image representation.
Geometrically, this implies allowing arbitrary signal distortions
in an $n$-dimensional rectangle defined by the
$\varepsilon$-insensitivity zone in each dimension of the selected
image representation domain. Unfortunately, not every image
representation domain is well-suited for such a simple,
scalar-wise, approach because statistical and/or perceptual
interactions between the coefficients may exist. These
interactions imply that scalar approaches may induce distortions
that do not follow the image statistics and/or are perceptually
annoying. Taking into account these relations would imply using
non-rectangular $\varepsilon$-insensitivity regions (allowing
coupled distortions in different coefficients), which is beyond
the conventional SVM formulation.

In this paper, we report a condition on the suitable domain for
developing efficient SVM image coding schemes.
We analytically demonstrate that no linear domain fulfills this
condition because of the statistical and perceptual
inter-coefficient relations that exist in these domains.
This theoretical result is experimentally
confirmed by comparing SVM learning in previously reported linear
domains and in a recently proposed non-linear perceptual domain
that simultaneously reduces the statistical and perceptual
relations (so it is closer to fulfilling the proposed condition).
These results highlight the relevance of an appropriate choice of
the image representation before SVM learning.
\end{abstract}

\begin{keywords}
image coding, non-linear perception models, statistical
independence, support vector machines, insensitivity zone
\end{keywords}

\section{Problem Statement: The Diagonal Jacobian Condition}
\label{statement}

Image coding schemes based on support vector machines (SVM) have
been successfully introduced in the literature. SVMs have been
used in the spatial domain \citep{Robinson00}, in the block-DCT
domain \citep{Robinson03}, and in the wavelet
domain \citep{Ahmed05,Jiao05}. These coding methods take advantage
of the ability of the support vector regression (SVR) algorithm
for function approximation using a small number of parameters
(signal samples, or support vectors) \citep{Smola04}. In all
current SVM-based image coding techniques, a representation of the
image is described by the entropy-coded weights associated to the
support vectors necessary to approximate the signal with a given
accuracy. Relaxing the accuracy bounds reduces the number of
needed support vectors. In a given representation domain, reducing
the number of support vectors increases the compression ratio at
the expense of bigger distortion (lower image quality). By
applying the standard SVR formulation, a certain amount of
distortion in each sample of the image representation is allowed.
In the original formulation, scalar restrictions on the errors are
introduced using a constant $\varepsilon$-insensitivity value for
every sample.

Recently, this procedure has been refined by \citet{Gomez05} 
using a profile-dependent SVR \citep{Camps01NIPS} that considers a
different $\varepsilon$ for each sample or frequency. This
frequency-dependent insensitivity, $\varepsilon_f$, accounts for
the fact that, according to simple (linear) perception models, not
every sample in linear frequency domains (such as DCT or wavelets)
contributes to the perceived distortion in the same way.

Despite different domains have been proposed for SVM training
(spatial domain, block-DCT and wavelets) and different
$\varepsilon$ insensitivities per sample have been proposed, in
conventional SVR formulation, the particular distortions
introduced by regression in the different samples are not coupled.
In all the reported SVM-based image coding schemes, the RBF kernel
is used and the penalization parameter is fixed to an arbitrarily
large value. In this setting, considering $n$-sample signals as
$n$-dimensional vectors, the SVR guarantees that the approximated
vectors are confined in $n$-dimensional rectangles around the
original vectors. These rectangles are just $n$-dimensional cubes
in the standard formulation or they have certain elongation if
different $\varepsilon_f$ are considered in each axis, $f$.
Therefore, in all the reported SVM-based coding methods, these
rectangles are always oriented along the axes of the (linear)
image representation. According to this, a common feature of these
(scalar-wise) approaches is that they give rise to decoupled
distortions in each dimension. \citet{Perez02icann} proposed a
hyperspherical insensitivity zone to correct the
penalization factor in each dimension of multi-output regression
problems, but again, restrictions to each sample were still
uncoupled.

This scalar-wise strategy is not the best option in domains where
the different dimensions of the image representation are not
independent. For instance, consider the situation where actually
independent components, $\mathbf{r}_f$, are obtained from a given
image representation, $\mathbf{y}$, applying some eventually
non-linear transform, $R$:
\begin{equation}
\mathbf{y} \stackrel{R}{\longrightarrow} \mathbf{r}.
\nonumber
\label{global_model}
\end{equation}
In this case, SVM regression with scalar-wise error restriction
makes sense in the $\mathbf{r}$ domain. However, the original
$\mathbf{y}$ domain will not be suitable for the standard SVM
regression unless the matrix $\nabla R$ is diagonal (up to any
permutation of the dimensions, that is, only one non-zero element
per row). Therefore, if transforms that achieve independence have
non-diagonal Jacobian, scalar-wise restrictions in the original
(coupled coefficients) domain $\mathbf{y}$ are not allowed.

\begin{figure}
\hspace{-0.25cm} \centerline{
\IG[width=0.9\textwidth]{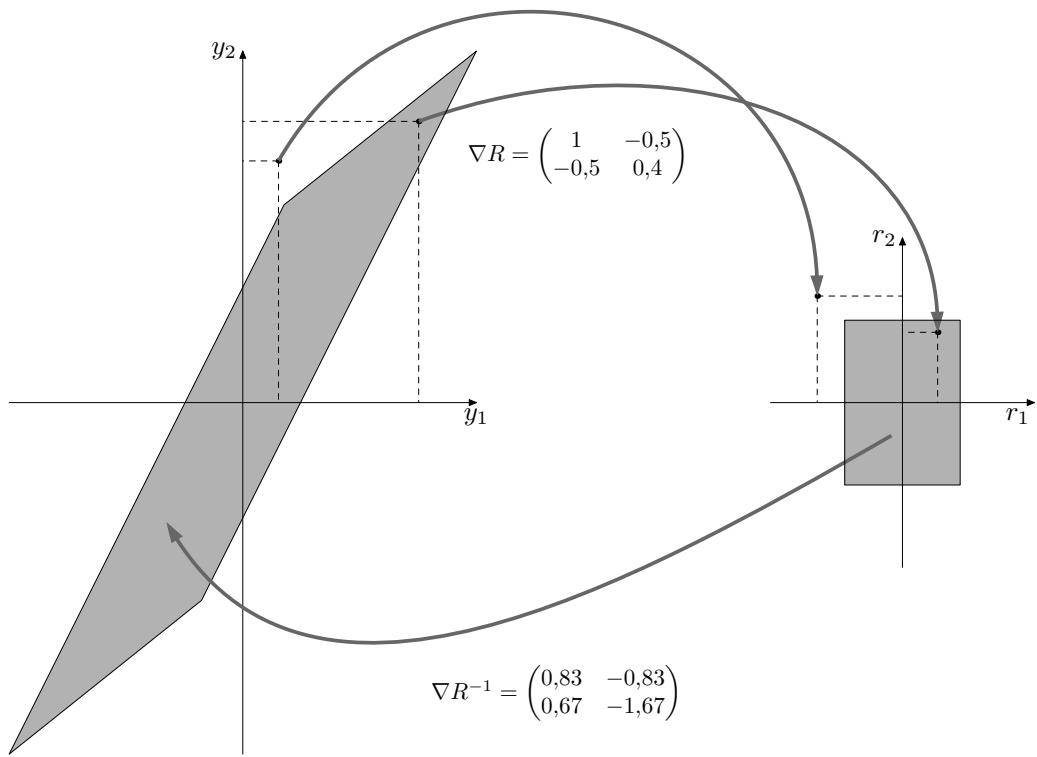}}
\caption{Insensitivity regions in different representation
domains, \textbf{y} (left) and \textbf{r} (right), related by a
non-diagonal transform $\nabla R$ and its inverse $\nabla R^{-1}$.}
\label{caja_2D}
\end{figure}

Figure~\ref{caja_2D} illustrates this situation. The shaded region
in the right plot ($\mathbf{r}$ domain) represents the
$n$-dimensional box determined by the $\varepsilon_f$
insensitivities in each dimension ($f$=1,2), in which a
scalar-wise approach is appropriate due to independence among
signal coefficients. Given that the particular $\nabla R$
transform is not diagonal, the corresponding shaded region in the
left plot (the original $\mathbf{y}$ domain) is not aligned along
the axes of the representation. This has negative implications:
note that for the highlighted points, smaller distortions in both
dimensions in the $\mathbf{y}$ domain (as implied by SVM with
tighter \emph{but scalar} $\varepsilon_f$ insensitivities) do not
necessarily imply lying inside the insensitivity region in the
final truly independent (and meaningful) $\mathbf{r}$ domain.
Therefore, the original $\mathbf{y}$ domain is not suitable for
the direct application of conventional SVM, and consequently,
non-trivial coupled insensitivity regions are required.

Summarizing, in the image coding context, the condition for an
image representation \textbf{y} to be strictly suitable for
conventional SVM learning is that \emph{the transform that maps
the original representation} \textbf{y} \emph{to an independent
coefficient representation} \textbf{r} \emph{must be locally
diagonal}.

As will be reviewed below, independence among coefficients (and
the transforms to obtain them) may be defined in both statistical
and perceptual terms \citep{Oja01,Malo01a,Epifanio03,Malo06a}. On
the one hand, a locally diagonal relation to a statistically
independent representation is desirable because independently
induced distortions (as the conventional SVM approach does) will
preserve the statistics of the distorted signal, that is, it will not
introduce artificial-looking artifacts. On the other hand, a
locally diagonal relation to a perceptually independent
representation is desirable because independently induced
distortions do not give rise to increased subjective distortions
due to non-trivial masking or facilitation interactions between
the distortions in each dimension \citep{Watson97a}.

In this work, we show that conventional linear domains do not
fulfill the diagonal Jacobian condition in either the statistical
case or in the perceptual case. This theoretical result is
experimentally confirmed by comparing SVM learning in previously
reported linear domains \citep{Robinson03,Gomez05} and in a
recently proposed non-linear perceptual domain that simultaneously
reduces the statistical and the perceptual
relations \citep{Malo06a}, thus, this non-linear perceptual domain
is closer to fulfilling the proposed condition.

The rest of the paper is structured as follows.
Section~\ref{relations} reviews the fact that linear coefficients
of the image representations commonly used for SVM training are
neither statistically independent nor perceptually independent.
Section~\ref{non_diagonal} shows that transforms for obtaining
statistical and/or perceptual independence from linear domains
have non-diagonal Jacobian. This suggests that there is room to
improve the performance of conventional SVM learning reported in
linear domains. In Section~\ref{alternative}, we propose the use of a perceptual
representation for SVM training because it strictly fulfills the
diagonal Jacobian condition in the perceptual sense and increases
the statistical independence among coefficients, bringing it
closer to fulfilling the condition in the statistical sense. The
experimental image coding results confirm the superiority of this
domain for SVM training in Section~\ref{performance}. Section
\ref{conclusions} presents the conclusions and final remarks.

\section{Statistical and Perceptual Relations Among Image Coefficients}
\label{relations}

Statistical independence among the coefficients of a signal
representation refers to the fact that the joint PDF of the class
of signals to be considered can be expressed as a product of the
marginal PDFs in each dimension \citep{Oja01}. Simple
(second-order) descriptions of statistical dependence use the
non-diagonal nature of the covariance
matrix \citep{Clarke85,Gersho92}. More recent and accurate
descriptions use higher-order moments, mutual information, or the
non-Gaussian nature (sparsity) of marginal
PDFs \citep{Oja01,Simoncelli97}.

Perceptual independence refers to the fact that the visibility of
errors in coefficients of an image may depend on the energy of
neighboring coefficients, a phenomenon known in the perceptual
literature as {\em masking} or {\em
facilitation} \citep{Watson97a}. Perceptual dependence has been
formalized just up to second order, and this may be described by
the non-Euclidean nature of the perceptual metric
matrix \citep{Malo01a,Epifanio03,Malo06a}.

\subsection{Statistical Relations}

In recent years, a variety of approaches, known collectively as
``independent component analysis'' (ICA), have been developed to
exploit higher-order statistics for the purpose of achieving a
unique \emph{linear} solution for coefficient
independence \citep{Oja01}. The basis functions obtained when
these methods are applied to images are spatially localized and
selective for both orientation and spatial
frequency \citep{Olshausen96,Bell97}. Thus, they are similar to
basis functions of multi-scale wavelet representations.

Despite its name, linear ICA does {\em not} actually produce
statistically independent coefficients when applied to
photographic images. Intuitively, independence would seem
unlikely, since images are not formed from linear superpositions
of independent patterns: the typical combination rule for the
elements of an image is {\em occlusion}. Empirically, the
coefficients of natural image decompositions in spatially
localized oscillating basis functions are found to be fairly well
decorrelated (i.e., their covariance is almost zero). However, the
amplitudes of coefficients at nearby spatial positions,
orientations, and scales are highly correlated (even with
orthonormal
transforms) \citep{Simoncelli97,Buccigrossi99,Wainwright01a,Hyvarinen03,Gutierrez06,Malo06a,Malo06b}.
This suggests that achieving statistical independence requires the
introduction of non-linearities beyond linear ICA transforms.

\begin{figure}
   \centering
   \begin{tabular}{cc}
     \IG[width=0.45\textwidth]{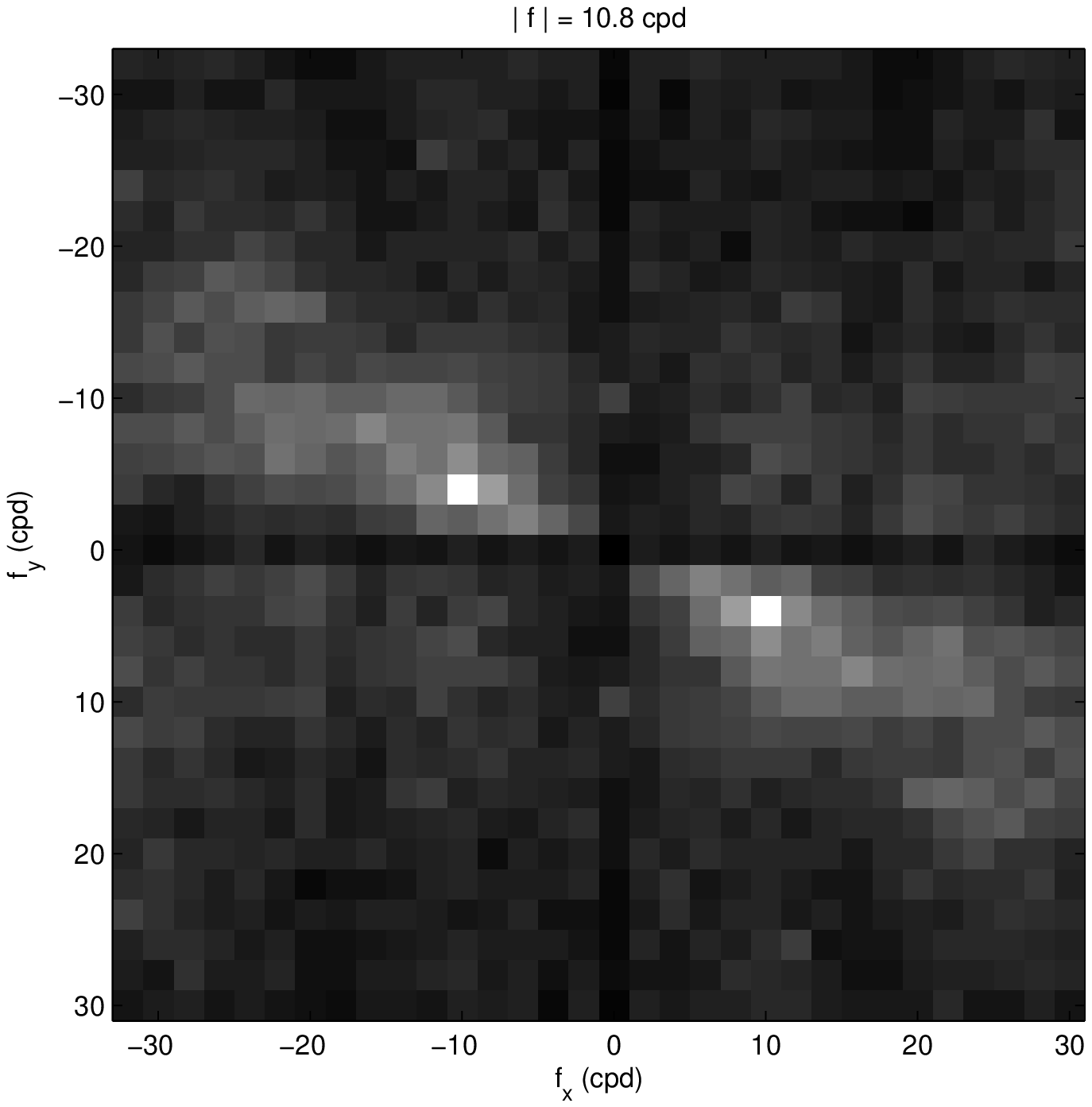}     &
     \IG[width=0.45\textwidth]{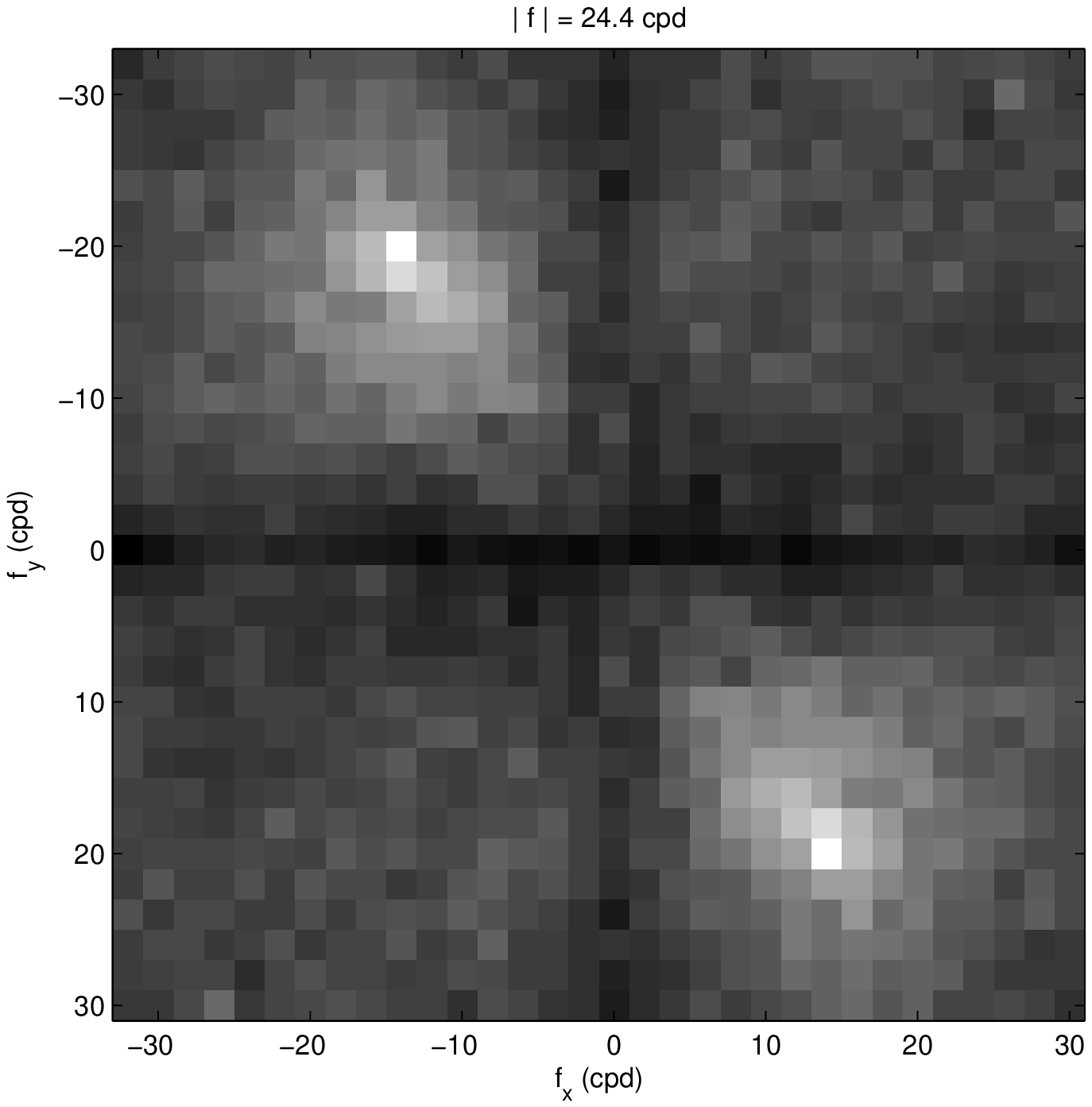} \\
   \end{tabular}
  \caption{Statistical interaction of two particular coefficients of the local Fourier Transform
  with their neighbors in a natural image database. The absolute value of the frequency of these coefficients
  is $\vert f \vert = 10.8 $ and $\vert f \vert = 24.4 $ cycles/degree (cpd). }
  \label{gausianas}
\end{figure}

Figure \ref{gausianas} reproduces one of many results that
highlight the presence of statistical relations of natural image
coefficients in block PCA or linear ICA-like domains: the energy
of spatially localized oscillating filters is correlated with the
energy of neighboring filters in scale and orientation 
\citep[see][]{Gutierrez06}. A remarkable feature is that
the interaction width increases with frequency, as has been
reported in other domains, for example, wavelets \citep{Buccigrossi99,Wainwright01a,Hyvarinen03}, and
block-DCT \citep{Malo06a}.

In order to remove the remaining statistical relations in the
linear domains \textbf{y}, non-linear ICA methods are
necessary \citep{Oja01,Lin99,Karhunen00,Jutten03}. Without lack of
generality, non-linear ICA transforms can be schematically
understood as a two-stage process \citep{Malo06b}:
\begin{equation}
      \xymatrix{
      \mathbf{x} \ar@/^/[r]^{\mathbf{T}} \ar@/_/@{<-}[r]_{\mathbf{T^{-1}}} &
      \mathbf{y} \ar@/^/[r]^{R} \ar@/_/@{<-}[r]_{R^{-1}} & \mathbf{r}
      },
      \label{modelV1}
\end{equation}
\noindent where \textbf{x} is the image representation in the
spatial domain, and $\mathbf{T}$  is a global unitary linear
transform that removes second-order and eventually higher-order
relations among coefficients in the spatial domain. Particular
examples of $\mathbf{T}$ include block PCA, linear ICAs, DCT or
wavelets. In the ICA literature notation, $\mathbf{T}$ is the
\emph{separating matrix} and $\mathbf{T^{-1}}$ is the \emph{mixing
matrix}. The second transform $R$ is an additional non-linearity
that is introduced in order to remove the statistical relations
that still remain in the \textbf{y} domain.

\subsection{Perceptual Relations}\label{percepualrelations}

Perceptual dependence among coefficients in different image
representations can be understood by using the current model of V1
cortex. This model can also be summarized by the two-stage (linear
and non-linear) process described in Equation~\eqref{modelV1}. In this
perceptual case, $\mathbf{T}$ is also a linear filter bank applied
to the original input image in the spatial domain. This filter
bank represents the linear behavior of V1 neurons whose receptive
fields happen to be similar to wavelets or linear ICA basis
functions \citep{Olshausen96,Bell97}. The second transform, $R$,
is a non-linear function that accounts for the masking and
facilitation phenomena that have been reported in the linear
\textbf{y} domain \citep{Foley94,Watson97a}.
Section~\ref{non_diagonal_per} gives a parametric expression for
the second non-linear stage, $R$: the divisive normalization
model \citep{Heeger92,Foley94,Watson97a}.

This class of models is based on psychophysical experiments
assuming that the last domain, \textbf{r}, is perceptually
Euclidean (i.e., perfect perceptual independence). An additional
confirmation of this assumption is the success of (Euclidean)
subjective image distortion measures defined in that
domain \citep{Teo94b}. Straightforward application of Riemannian
geometry to obtain the perceptual metric matrix in other domains
shows that the coefficients of linear domains \textbf{x} and
\textbf{y}, or any other linear transform of them, are not
perceptually independent \citep{Epifanio03}.

\begin{figure}
   \begin{center}
    \begin{tabular}{cccc}
\hspace{-0.75cm}
      \small  3 cpd & \small 6 cpd & \small 12 cpd & \small 24 cpd\\
  \IG[width=0.24\textwidth]{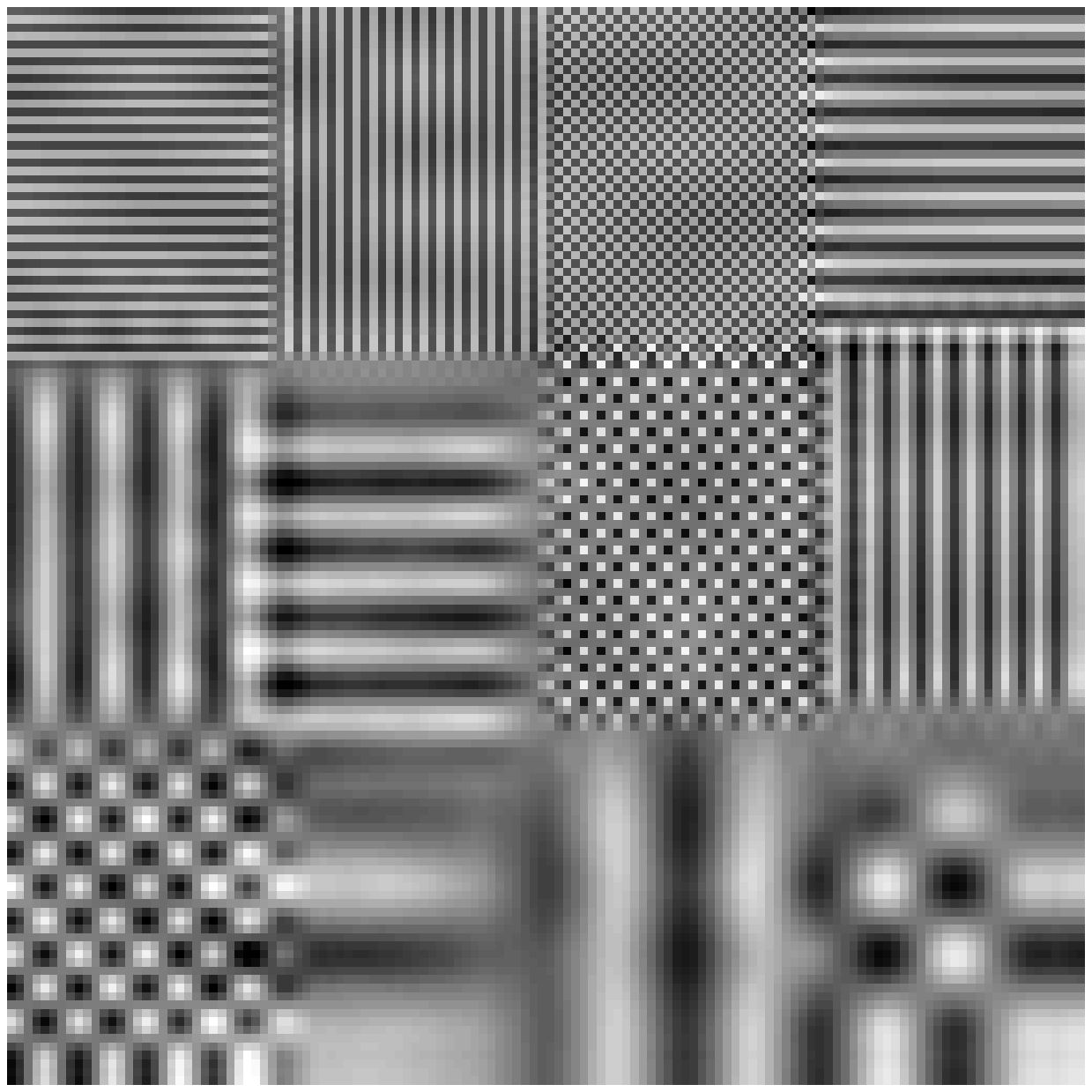} & \hspace{-0.35cm} \IG[width=0.24\textwidth]{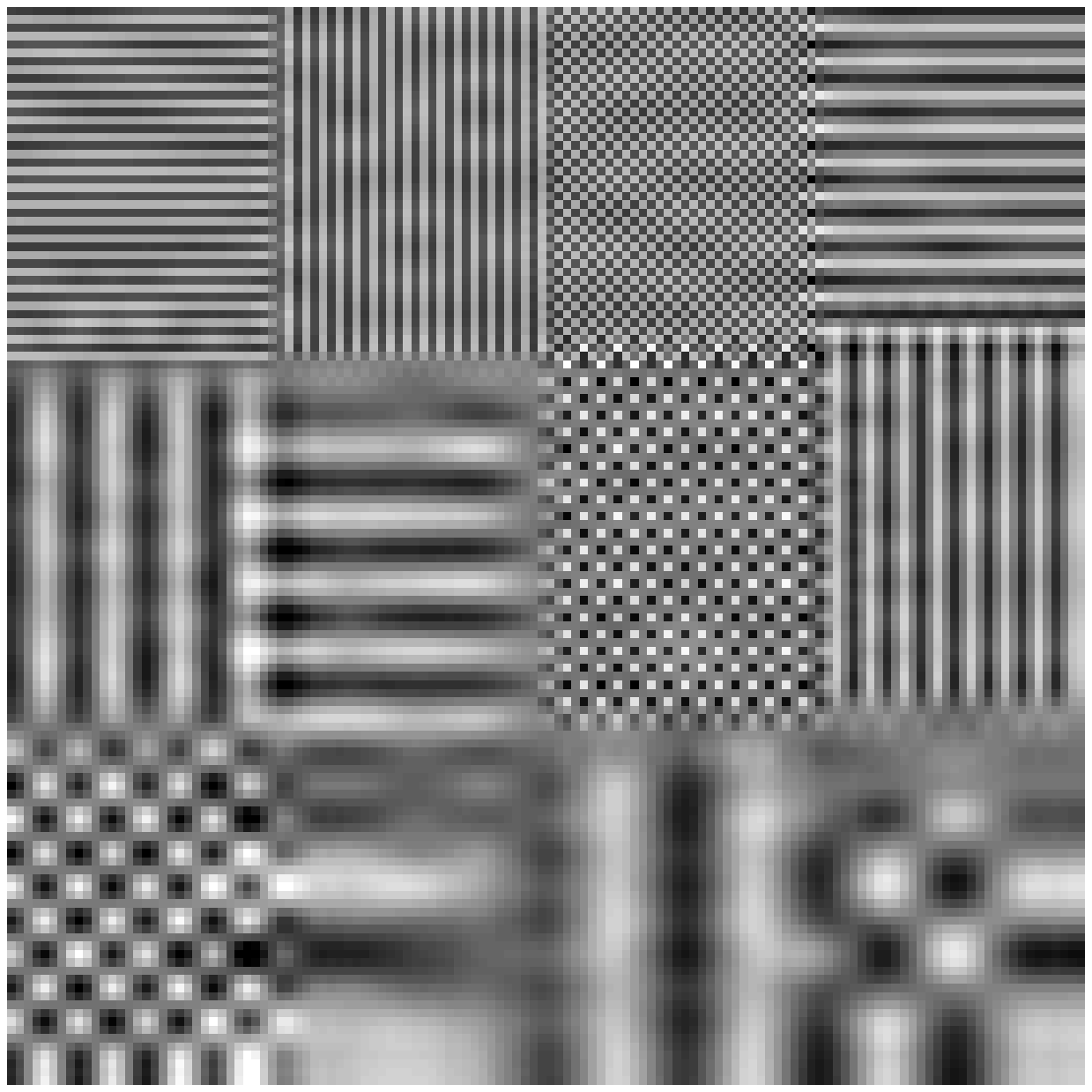} & \hspace{-0.35cm} \IG[width=0.24\textwidth]{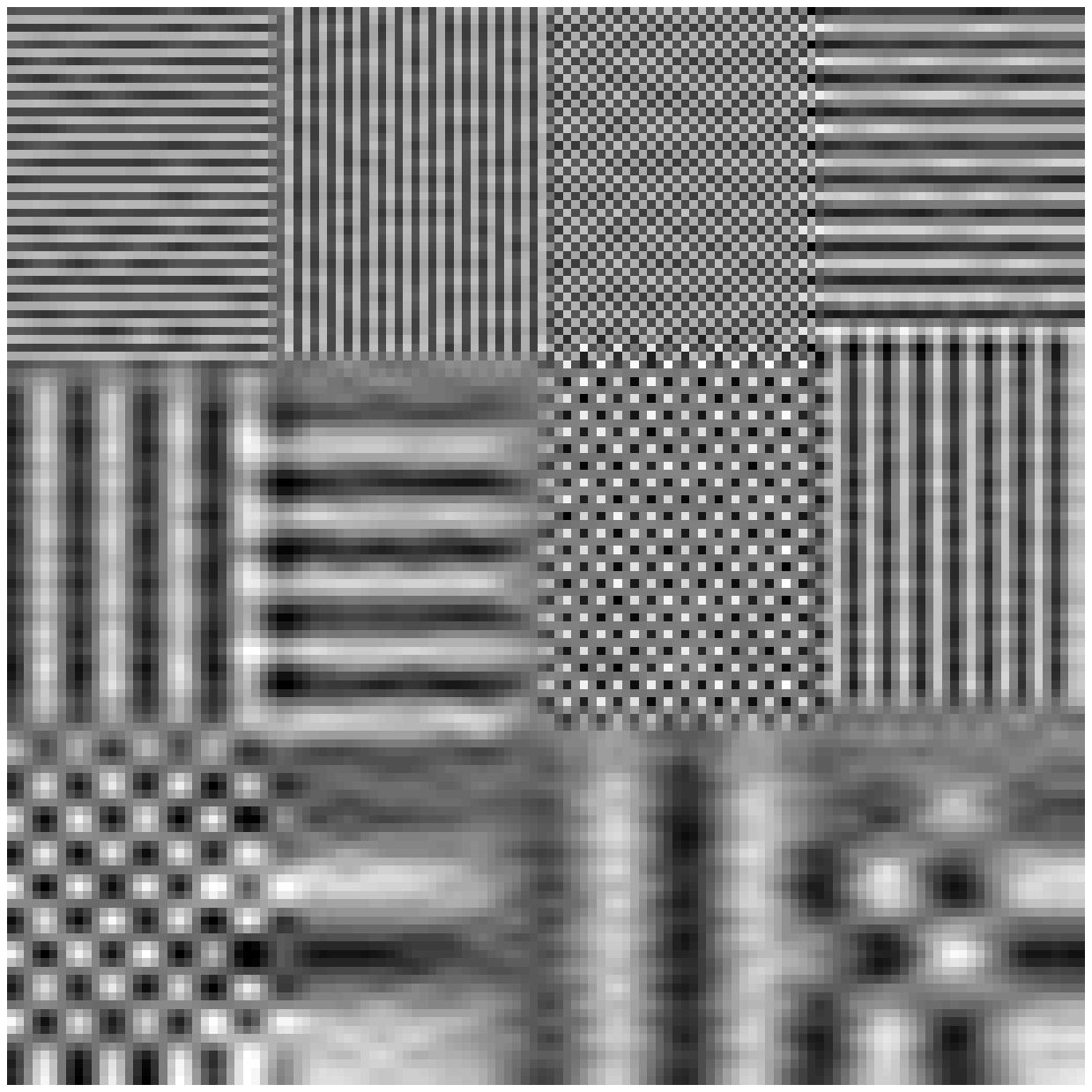}& \hspace{-0.35cm} \IG[width=0.24\textwidth]{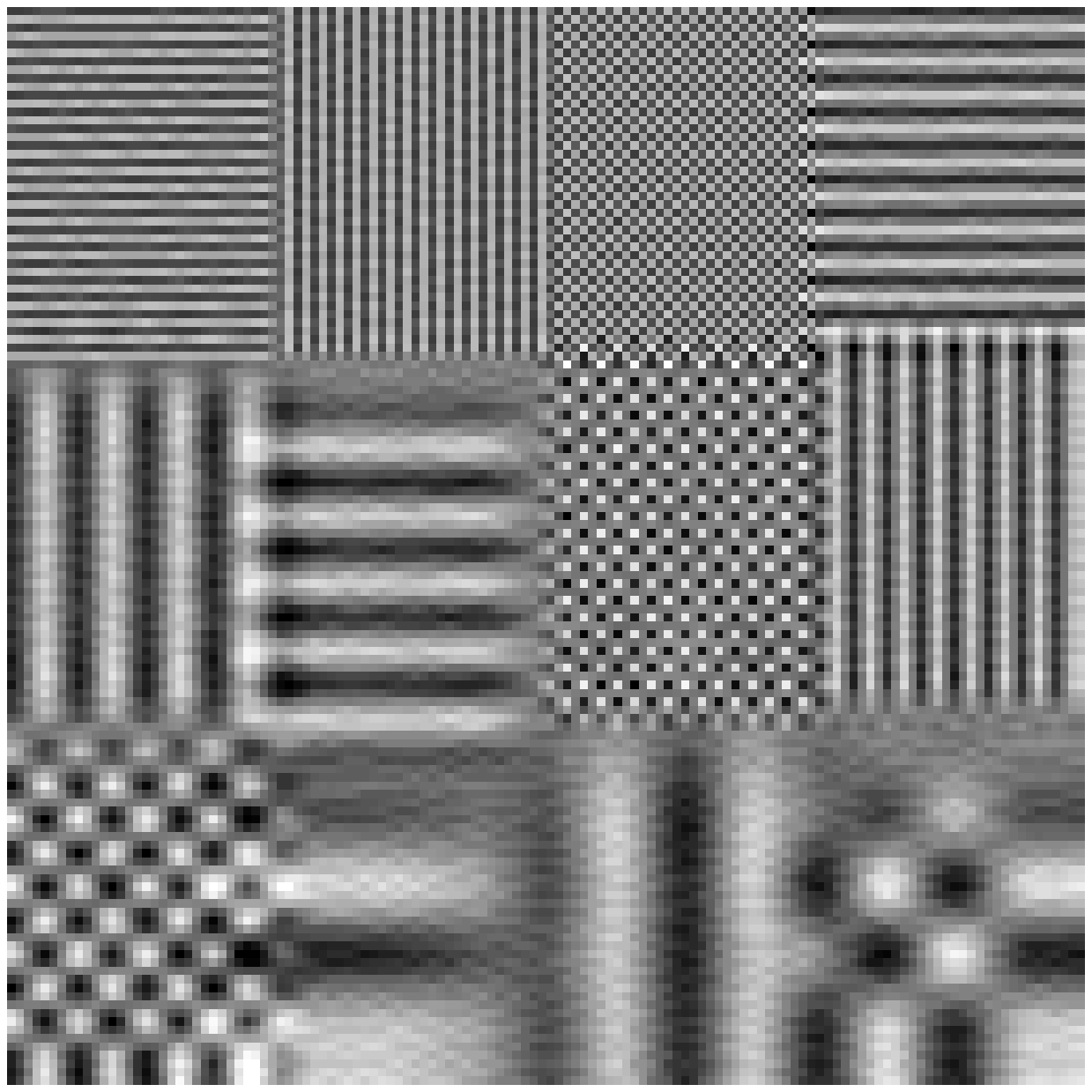}\\
   \end{tabular}
   \end{center}
   \caption{Illustrative example of perceptual dependence (cross-masking phenomenon). Equal energy noise of
   different frequency content, 3 cycl/deg (cpd), 6 cpd, 12 cpd and 24 cpd, shown on top
   of a background image. Sampling frequency assumes that these images subtend an angle of 3 deg.}
   \label{ruidos}
\end{figure}

Figure \ref{ruidos} illustrates the presence of perceptual
relations between coefficients when using linear block frequency
or wavelet-like domains, \textbf{y}: the {\em cross-masking}
behavior. In this example, the visibility of the distortions added
on top of the background image made of periodic patterns has to be
assessed. This is a measure of the sensitivity of a particular
perceptual mechanism to distortions in that dimension, $\Delta
y_f$, when mechanisms tuned to other dimensions are simultaneously
active, that is, $y_{f'}\neq 0$, with $f' \neq f$. As can be
observed, low frequency noise is more visible in high frequency
backgrounds than in low frequency backgrounds (e.g., left image).
Similarly, high frequency noise is more visible in low frequency
backgrounds than in high frequency ones (e.g., right image). That
is to say, a signal of a specific frequency strongly masks the
corresponding frequency analyzer, but it induces a smaller
sensitivity reduction in the analyzers that are tuned to different
frequencies. In other words, the reduction in sensitivity of a
specific analyzer gets larger as the distance between the
background frequency and the frequency of the analyzer gets
smaller. The response of each frequency analyzer not only depends
on the energy of the signal for that frequency band, but also on
the energy of the signal in other frequency bands (cross-masking).
This implies that a different amount of noise in each frequency
band may be acceptable depending on the energy of that frequency
band and on the energy of neighboring bands. This is what we have
called {\em perceptual dependence} among different coefficients in
the \textbf{y} domain.

At this point, it is important to stress the similarity between the
set of computations to obtain statistically decoupled image
coefficients and the known stages of biological vision. In fact,
it has been hypothesized that biological visual systems have
organized their sensors to exploit the particular statistics of
the signals they have to process.
See \citet{Barlow01}, \citet{Simoncelli01}, and \citet{Simoncelli03} for reviews on this
hypothesis.

In particular, both the linear and the non-linear stages of the
cortical processing have been successfully derived using
redundancy reduction arguments: nowadays, the same class of linear
stage \textbf{T} is used in transform coding algorithms and in
vision models \citep{Olshausen96,Bell97,Taubman01}, and new
evidence supports the same idea for the second non-linear
stage \citep{Schwartz01,Malo06b}. According to this, the
statistical and perceptual transforms, $R$, that remove the above
relations from the linear domains, \textbf{y}, would be very
similar if not the same.

\section{Statistical and Perceptual Independence Imply Non-diagonal Jacobian}
\label{non_diagonal}

In this section, we show that both statistical redundancy
reduction transforms (e.g., non-linear ICA) and perceptual
independence transforms (e.g., divisive normalization), have
non-diagonal Jacobian for any linear image representation, so they
are not strictly suitable for conventional SVM training.

\subsection{Non-diagonal Jacobian in Non-linear ICA Transforms}

One possible approach for dealing with global non-linear ICA is to
act differentially by breaking the problem into local linear
pieces that can then be integrated to obtain the global
independent coefficient domain \citep{Malo06b}. Each differential
sub-problem around a particular point (image) can be locally
solved using the standard linear ICA methods restricted to the
neighbors of that point \citep{Lin99}.

Using the differential approach in the context of a two-stage
process such as the one in Equation~\eqref{modelV1}, it can be shown
that \citep{Malo06b}:
\begin{equation}
      \mathbf{r} = \mathbf{r_0} + \int_{\mathbf{x}_0}^{\mathbf{x}} \mathbf{T}_{\boldsymbol{\ell}}(\mathbf{x'}) \, d\mathbf{x'}
      = \mathbf{r_0} + \int_{\mathbf{x}_0}^{\mathbf{x}} \nabla R(\mathbf{T x'}) \, \mathbf{T} \, d\mathbf{x'},
      \label{local_to_global_ICA}
\end{equation}
where $\mathbf{T}_{\boldsymbol{\ell}}(\mathbf{x'})$ is the local
separating matrix for a neighborhood of the image $\mathbf{x'}$,
and \textbf{T} is the global separating matrix for the whole PDF.
Therefore, the Jacobian of the second non-linear stage is:
\begin{equation}
      \nabla R(\mathbf{y}) = \nabla R(\mathbf{T} \mathbf{x}) =\mathbf{T}_{\boldsymbol{\ell}}(\mathbf{x}) \, \mathbf{T^{-1}}.
      \label{JacobianV1}
\end{equation}

As local linear independent features around a particular image,
\textbf{x}, differ in general from global linear independent
features, that is, $\mathbf{T}_{\boldsymbol{\ell}}(\mathbf{x}) \neq
\mathbf{T}$, the above product is not the identity nor diagonal in
general.

\subsection{Non-diagonal Jacobian in Non-linear Perceptual Transforms}
\label{non_diagonal_per}

\begin{figure}
   \begin{center}
    \begin{tabular}{cc}
        \IG[width=0.35\textwidth]{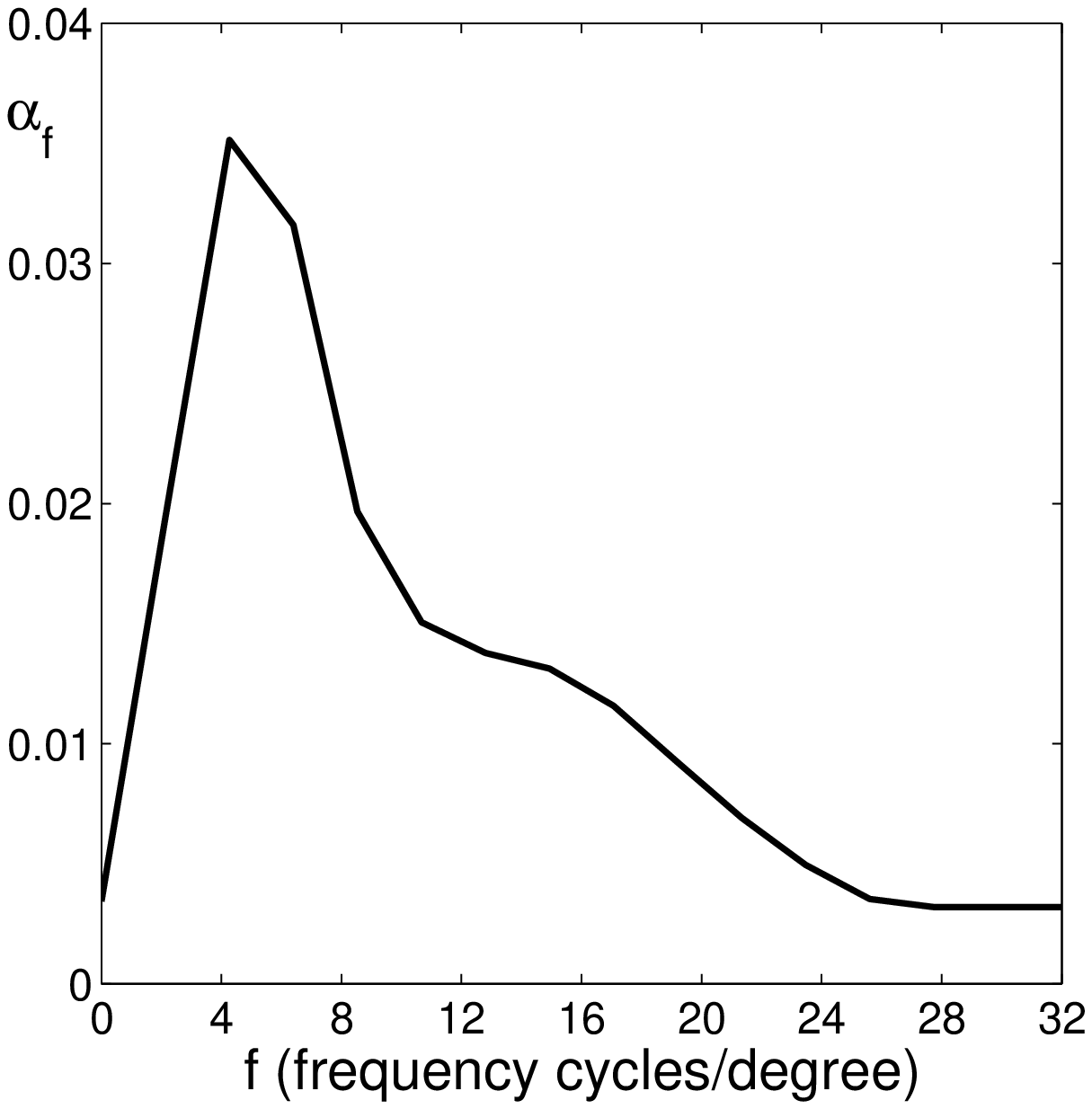} &
        \IG[width=0.35\textwidth]{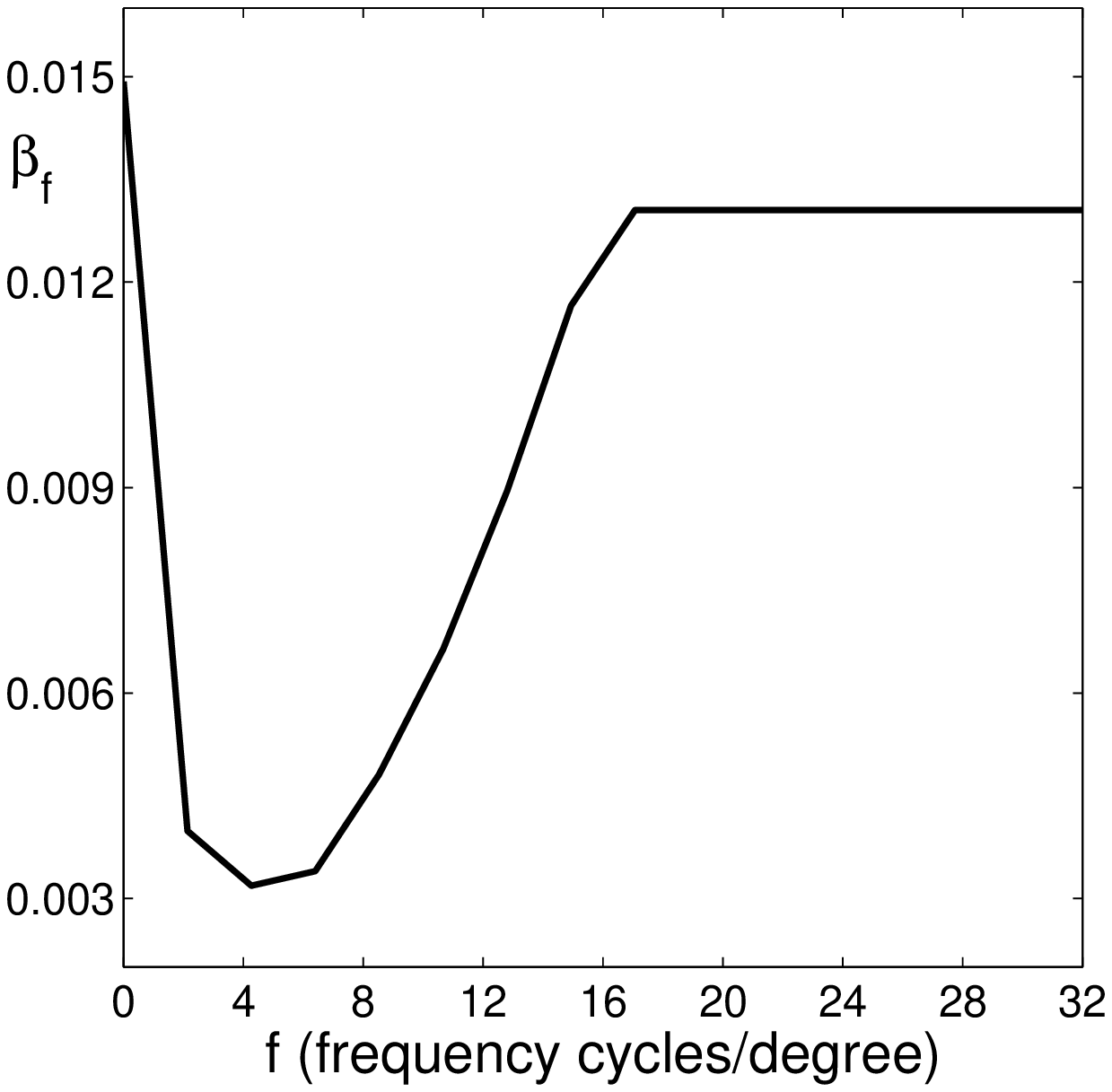} \\
        \IG[width=0.37\textwidth]{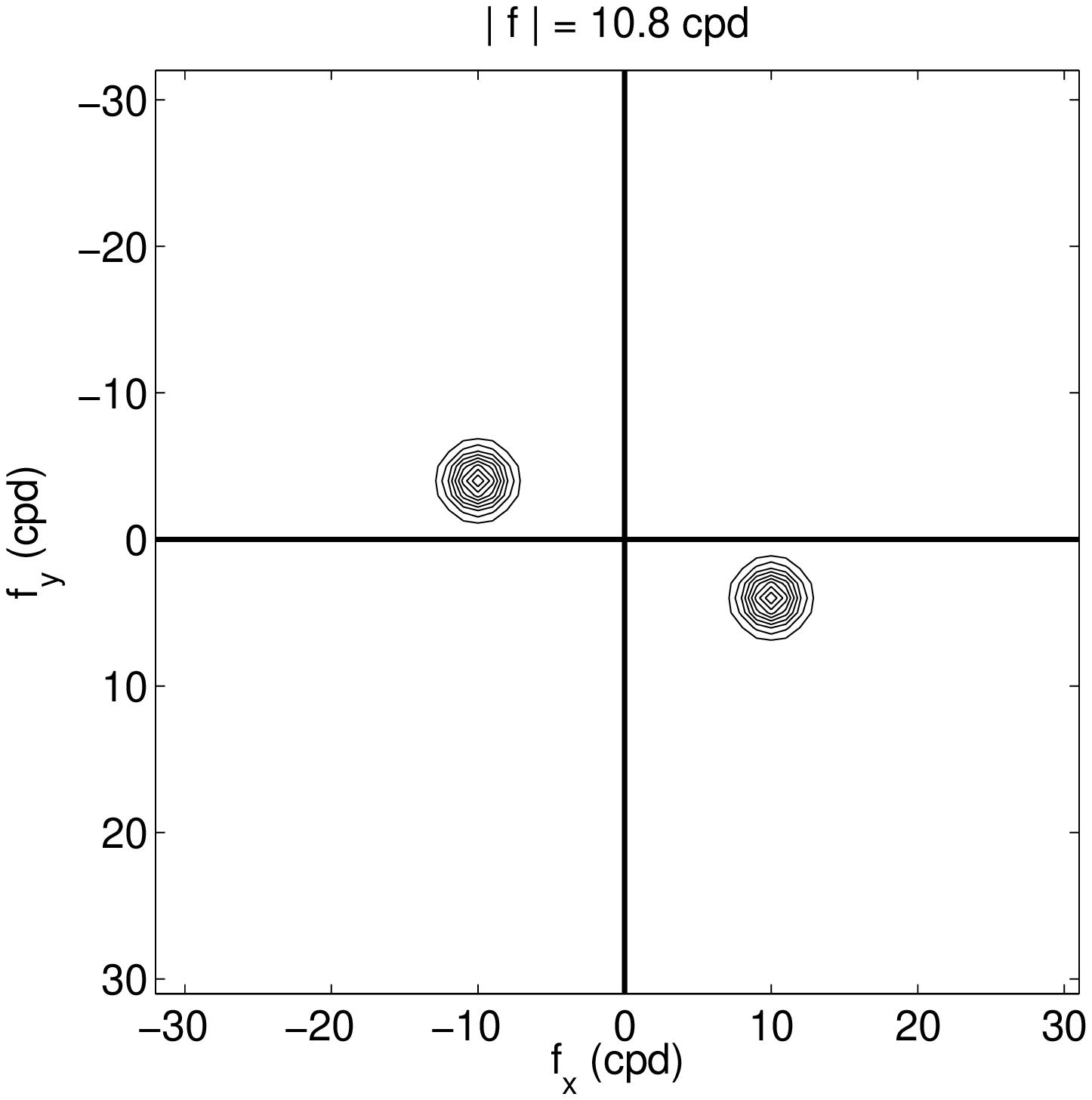} &
        \IG[width=0.37\textwidth]{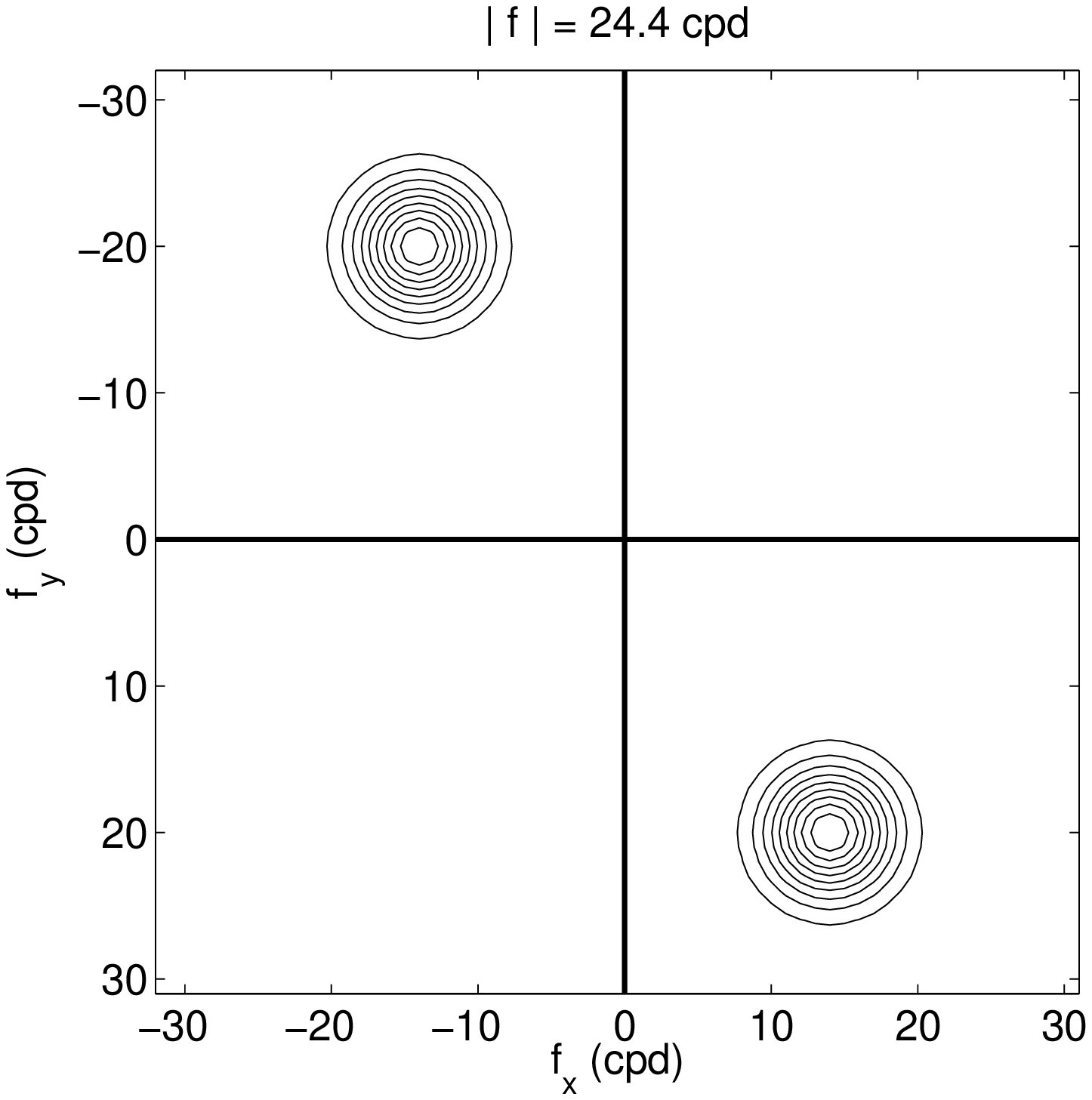} \\
   \end{tabular}
   \end{center}
\caption{Parameters of the perceptual model: $\alpha_f$ (top left), $\beta_f$ (top right).
Bottom figures represent perceptual interaction neighborhoods $h_{ff'}$ of two
particular coefficients of the local Fourier
domain.}
\label{non_linear_parameters}
\end{figure}

The current response model for the cortical frequency analyzers is
non-linear \citep{Heeger92,Watson97a}. The outputs of the filters
of the first linear stage, \textbf{y}, undergo a non-linear
sigmoid transform in which the energy of each linear coefficient
is weighted by a linear \emph{Contrast Sensitivity Function}
(CSF) \citep{Campbell68,Malo97a} and is further normalized by a
combination of the energies of neighbor coefficients in frequency,
\begin{equation}
r_{f} = R(\mathbf{y})_{f} = \frac{\sgn(y_f) \, |\alpha_{f} \,
y_{f}|^{\gamma}}{\beta_{f}+\sum_{f'=1}^{n} h_{ff'} |\alpha_{f'} \, y_{f'}|^{\gamma}},
\label{non_linear_model}
\end{equation}
where $\alpha_f$ (Figure~\ref{non_linear_parameters}[top left]) are
CSF-like weights, $\beta_f$ (Figure~\ref{non_linear_parameters}[top
right]) control the sharpness of the response saturation for each
coefficient, $\gamma$ is the so called excitation exponent, and
the matrix $h_{ff'}$ determines the interaction
neighborhood in the non-linear normalization of the energy.
This interaction matrix models the cross-masking behavior (cf. Section
\ref{percepualrelations}). The interaction in this matrix is
assumed to be Gaussian \citep{Watson97a}, and its width increases
with the frequency. Figure~\ref{non_linear_parameters}[bottom]
shows two examples of this Gaussian interaction for two particular
coefficients in a local Fourier domain. Note that the width of the
perceptual interaction neighborhood increases with the frequency
in the same way as the width of the statistical interaction
neighborhood shown in Figure~\ref{gausianas}. We used a value
of $\gamma=2$ in the experiments.

Taking derivatives in the general divisive normalization model,
Equation~\eqref{non_linear_model}, we obtain
\begin{equation}
\nabla R(\mathbf{y})_{ff'}= \sgn({y}_{f})\gamma\left(
\frac{ \alpha_f \, |\alpha_f \, y_f|^{\gamma-1}}{\beta_{f}+\sum_{f'=1}^{n} h_{ff'} |\alpha_{f'}
y_{{f'}}|^\gamma} \delta_{ff'} - \frac{ \alpha_{f'} |\alpha_{f} y_{f}|^{\gamma} |\alpha_{f'} y_{f'} |^{\gamma-1}}{(\beta_{f}+\sum_{f'=1}^{n} h_{ff'} |\alpha_{f'} y_{{f'}}|^\gamma)^2}  h_{ff'} \right),
\label{non_linear_jacobian}
\end{equation}
which is not diagonal because of the interaction matrix, $h$,
which describes the cross-masking between each frequency $f$ and
the remaining $f'\neq f$.

Note that the intrinsic non-linear nature of both the statistical
and perceptual transforms, Equations~\eqref{JacobianV1}
and~\eqref{non_linear_jacobian}, makes the above results true for
any linear domain under consideration. Specifically, if any other
possible linear domain for image representation is considered,
$\mathbf{y'}= \mathbf{T'} \, \mathbf{\mathbf{\mathbf{y}}}$, then
the Jacobian of the corresponding independence transform, $R'$, is
$$
      \nabla R'(\mathbf{y'}) = \nabla R(\mathbf{y}) \,
      \mathbf{T'^{-1}},
$$
which, in general, will also be non-diagonal because of the
non-diagonal and point-dependent nature of $\nabla R(\mathbf{y})$.

To summarize, since no linear domain fulfills the diagonal
Jacobian condition in either statistical or perceptual terms,
the negative situation illustrated in Figure~\ref{caja_2D} may occur
when using SVM in these domains. Therefore, improved results could
be obtained if SVM learning were applied after some transform
achieving independent coefficients, $R$.

\section{SVM Learning in a Perceptually Independent Representation}
\label{alternative}

In order to confirm the above theoretical results (i.e., the
unsuitability of linear representation domains for SVM learning)
and to assess the eventual gain that can be obtained from training
SVR in a more appropriate domain, we should compare the
performance of SVRs in previously reported linear domains
(e.g., block-DCT or wavelets) and in one of the proposed non-linear
domains (either the statistically independent domain or the
perceptually independent domain).

Exploration of the statistical independence transform may have
academic interest but, in its present formulation, it is not
practical for coding purposes: direct application of non-linear
ICA as in Equation~\eqref{local_to_global_ICA} is very time-consuming
for high dimensional vectors since lots of local ICA computations
are needed to transform each block, and a very large image
database is needed for a robust and significant computation of
$R$. Besides, an equally expensive differential approach is also
needed to compute the inverse $R^{-1}$ for image decoding. In
contrast, the perceptual non-linearity (and its inverse) are
analytical. These analytical expressions are feasible for
reasonable block sizes, and there are efficient iterative methods
that can be used for larger vectors \citep{Malo06a}. In this
paper, we explore the use of a psychophysically-based divisive
normalized domain: first compute a block-DCT transform and then
apply the divisive normalization model described above for each
block. The results will be compared to the first competitive SVM
coding results \citep{Robinson03} and the posterior improvements
reported by \citet{Gomez05}, both formulated in the linear
block-DCT domain.

As stated in Section~\ref{relations}, by construction, the
proposed domain is perceptually Euclidean with perceptually
independent components. The Euclidean nature of this domain has an
additional benefit: the $\varepsilon$-insensitivity design is very
simple because a constant value is appropriate due to the constant
perceptual relevance of all coefficients. Thus, direct application
of the standard SVR method is theoretically appropriate in this
domain.

Moreover, beyond its built-in perceptual benefits, this
psychophysically-based divisive normalization has attractive
statistical properties: it strongly reduces the mutual information
between the final coefficients \textbf{r} \citep{Malo06a}. This is
not surprising according to the hypothesis that try to explain the
early stages of biological vision systems using information theory
arguments \citep{Barlow61,Simoncelli01}. Specifically, dividing
the energy of each linear coefficient by the energy of the
neighbors, which are statistically related with it, cf.
Figure~\ref{gausianas}, gives coefficients with reduced statistical
dependence. Moreover, as the empirical non-linearities of
perception have been reproduced using non-linear ICA in
Equation~\eqref{local_to_global_ICA} \citep{Malo06b}, the empirical
divisive normalization can be seen as a convenient parametric way
to obtain statistical independence.

\section{Performance of SVM Learning in Different Domains}
\label{performance}

In this section, we analyze the performance of SVM-based coding
algorithms in linear and non-linear domains through
rate-distortion curves and explicit examples for visual
comparison. In addition, we discuss how SVM selects support
vectors in these domains to represent the image features.

\subsection{Model Development and Experimental Setup}

In the (linear) block-DCT domain, ${\bf y}$, we use the method
introduced by \citet{Robinson03} (RKi-1), in which the SVR is
trained to learn a fixed (low-pass) number of DCT coefficients
(those with frequency bigger than 20 cycl/deg are discarded); and
the method proposed by \citet{Gomez05} (CSF-SVR), in which the
relevance of all DCT coefficients is weighted according to the CSF
criterion using an appropriately modulated $\varepsilon_f$. In the
non-linear domain, ${\bf r}$, we use the SVR with constant
insensitivity parameter $\varepsilon$ (NL-SVR).
In all cases, the
block-size is 16$\times$16, that is, $\mathbf{y}, \, \mathbf{r} \in
\mathbb{R}^{256}$. The behavior of JPEG standard is also included in the experiments for
comparison purposes.

As stated in Section~\ref{statement}, we used the RBF kernel and
arbitrarily large penalization parameter in every SVR case. In all
experiments, we trained the SVR models without the bias term, and
modelled the absolute value of the DCT, \textbf{y}, or response
coefficients, \textbf{r}. All the remaining free parameters
($\varepsilon$-insensitivity and Gaussian width of the RBF kernel
$\sigma$) were optimized for all the considered models and
different compression ratios. In the NL-SVM case, the parameters
of the divisive normalization used in the experiments are shown in
Figure~\ref{non_linear_parameters}. After training, the signal is
described by the uniformly quantized Lagrange multipliers of the
support vectors needed to keep the regression error below the
thresholds $\varepsilon_f$. The last step is entropy coding of the
quantized weights. The compression ratio is controlled by a factor
applied to the thresholds, $\varepsilon_f$.

\subsection{Model Comparison}

\begin{figure}
\begin{center}
\begin{tabular}{ccc}
\IG[width=0.50\textwidth]{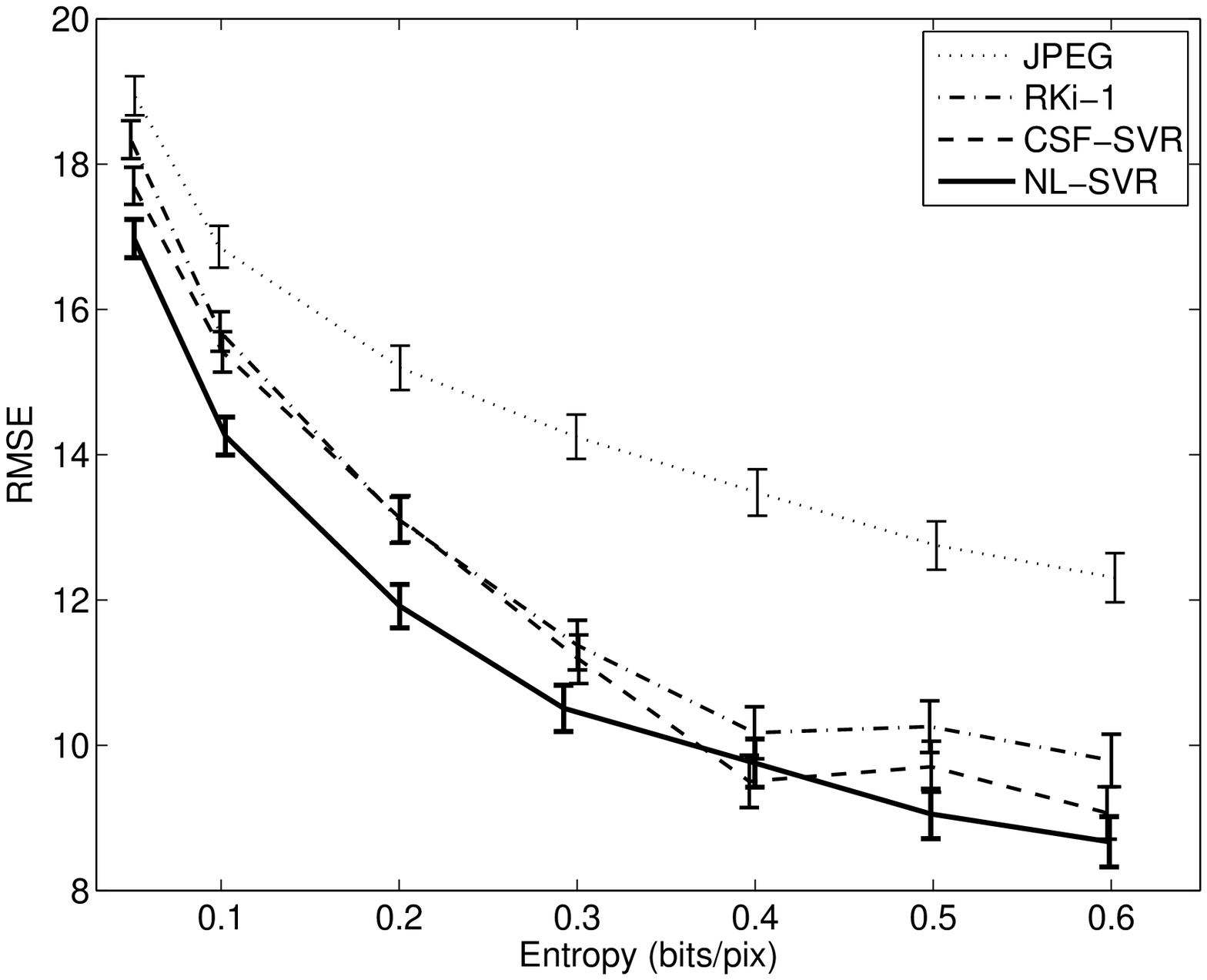}  \\
\IG[width=0.50\textwidth]{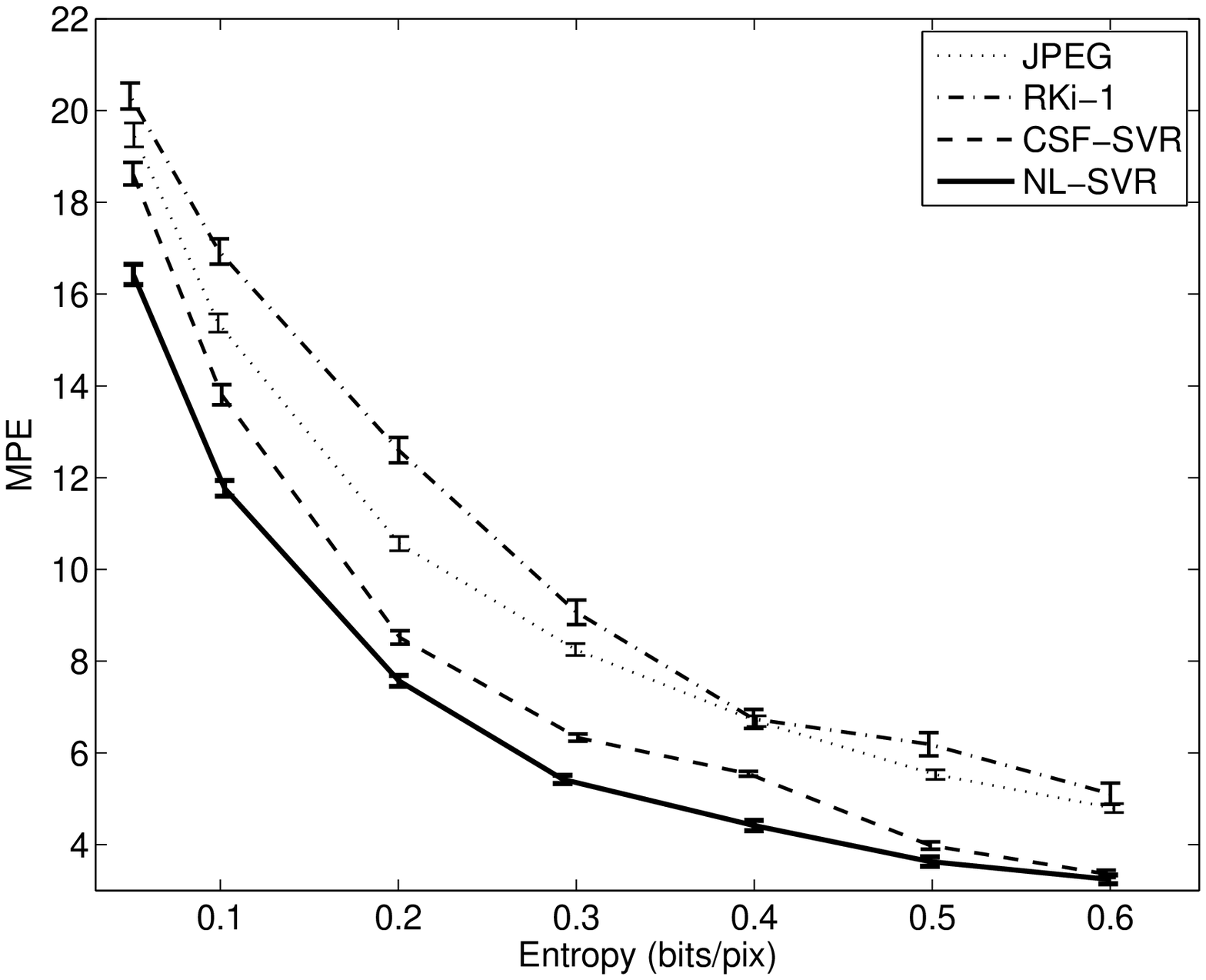}  \\
\IG[width=0.50\textwidth]{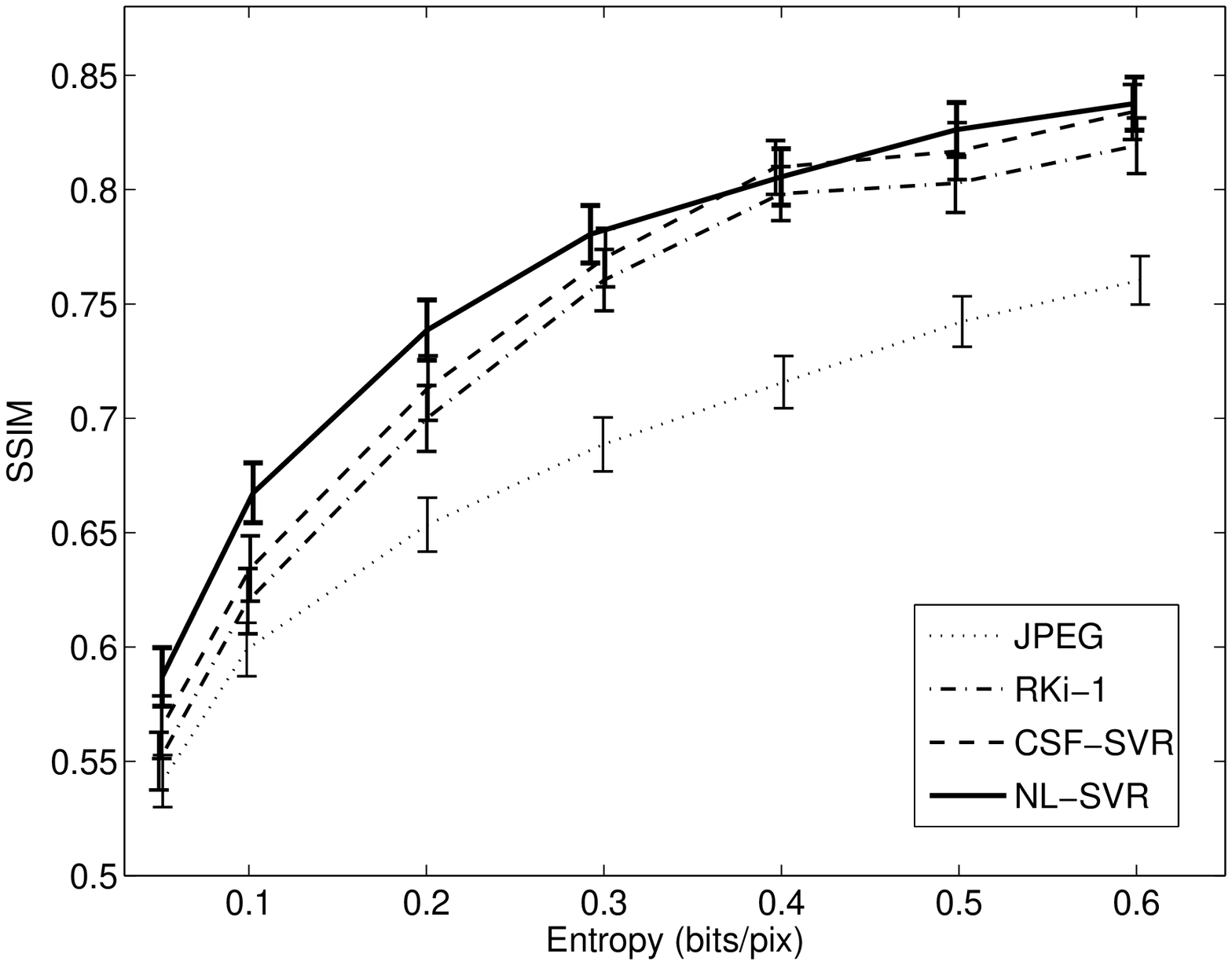}  \\
\end{tabular}
\end{center}
\vspace{-0.28in}
\caption{Average rate distortion curves over eight standard images
(Lena, Barbara, Boats, Einstein, Peppers, Mandrill, Goldhill, Camera man)
using objective and subjective
measures for the considered JPEG (dotted) and the SVM approaches
(RKi-1 dash-dotted, CSF-SVR dashed and NL-SVR solid). RMSE distortion (top),
Maximum Perceptual Error, MPE (middle) \citep{Malo00b,Gomez05,Malo06a}, and
Structural SIMilarity index, SSIM (bottom) \citep{Wang04}.} \label{rds}
\end{figure}

In order to assess the quality of the coded images, three
different measures were used: the standard (Euclidean) RMSE, the
Maximum Perceptual Error (MPE) \citep{Malo00b,Gomez05,Malo06a} and
the also perceptually meaningful Structural SIMilarity (SSIM)
index \citep{Wang04}. Eight standard 256$\times$256
monochrome 8 bits/pix images were used in the experiments.
Average rate-distortion curves are plotted in
Figure \ref{rds} in the range [0.05, 0.6] bits/pix (bpp).
According to these entropy-per-sample data, original file
size was 64 KBytes in every case, while the compressed image
sizes were in the range [0.4, 4.8] KBytes. This implies that
the compression ratios were in the range [160:1, 13:1].

In general, a clear gain
over standard JPEG is obtained by all SVM-based methods. According to the standard
Euclidean MSE point of view, the performance of RKi-1 and CSF-SVR
algorithms is basically the same (note the overlapped curves
in Figure~\ref{rds}(a)). However, it is widely known that the MSE
results are not useful to represent the subjective quality of
images, as extensively reported
elsewhere \citep{qual:Girod,Teo94b,Watson02}. When using more
appropriate (perceptually meaningful) quality measures
(Figures~\ref{rds}(b)-(c)), the CSF-SVR obtains a certain
advantage over the RKi-1 algorithm for all compression rates, which was
already reported by \citet{Gomez05}. In all measures, and for the
whole considered entropy range, the proposed NL-SVR clearly outperforms all
previously reported methods, obtaining a noticeable gain
at medium-to-high compression ratios (between 0.1 bpp (80:1)
and 0.3 bpp (27:1)). Taking into account that the
recommended bit rate for JPEG is about 0.5 bpp, from
Figure~\ref{rds} we can also conclude that the proposed technique achieves
the similar quality levels at a lower bit rate in the range [0.15, 0.3] bpp.

\begin{figure}
\begin{center}
\begin{tabular}{cc}
\IG[width=0.33\textwidth]{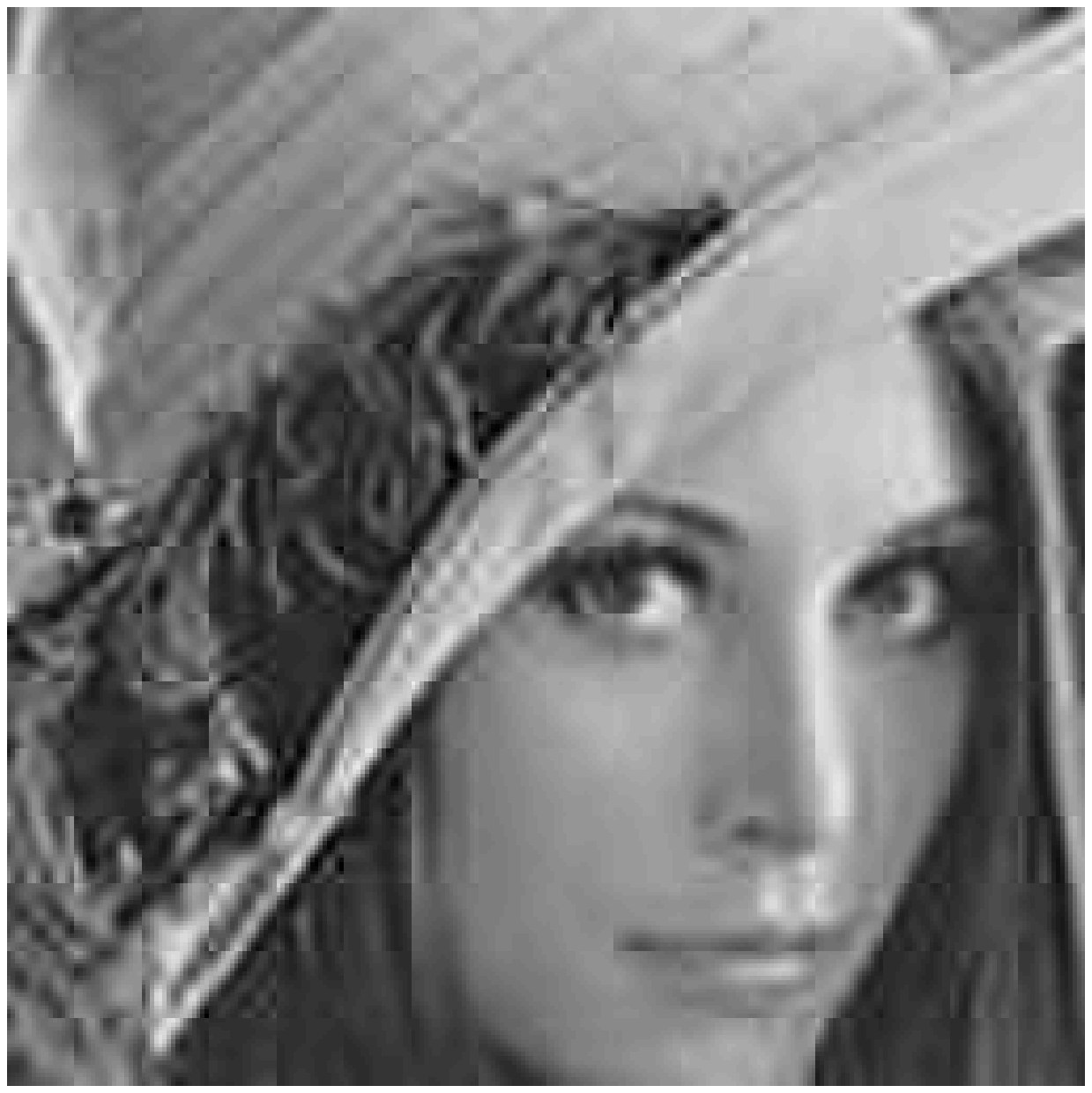} &  \IG[width=0.33\textwidth]{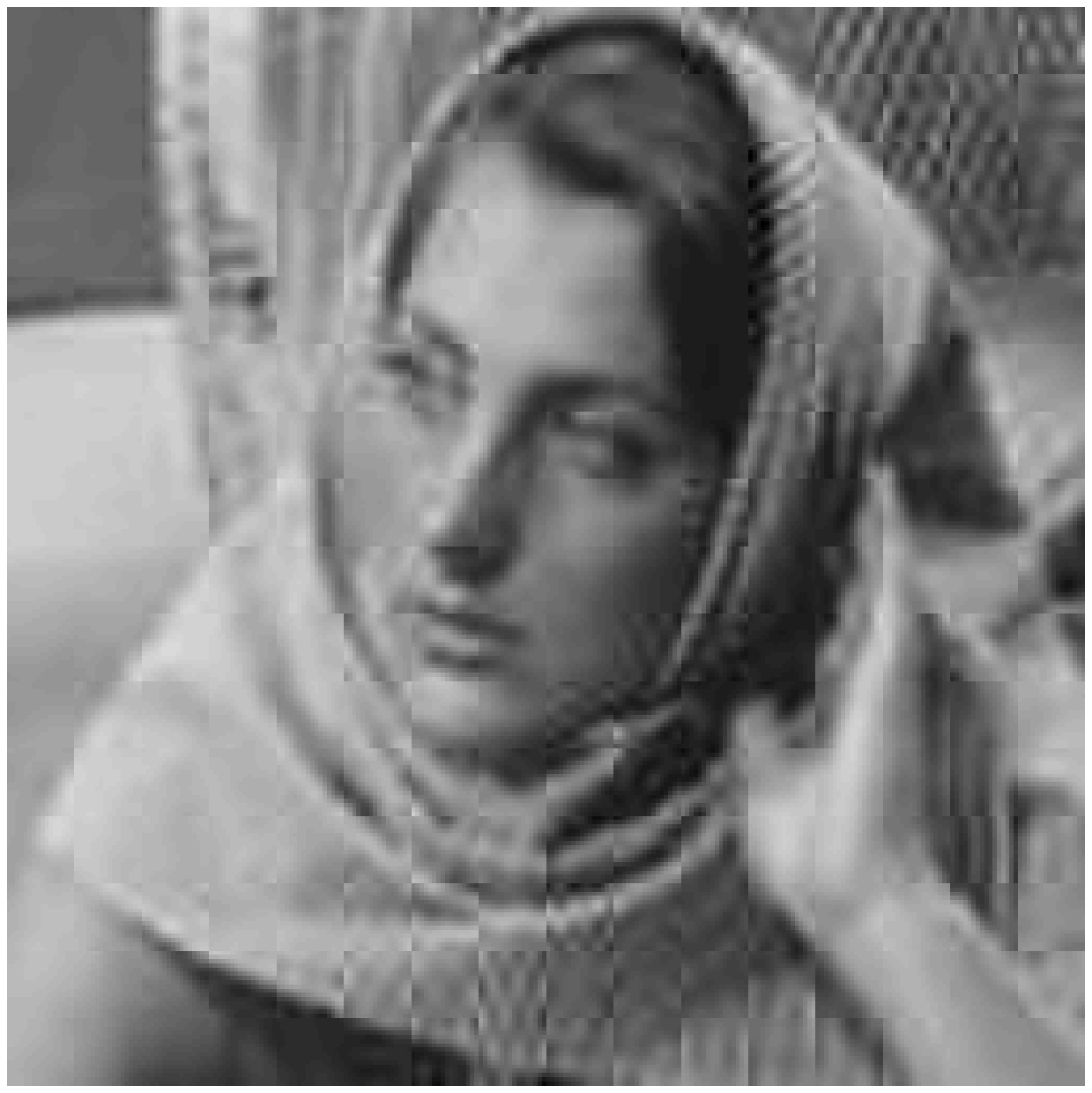} \\
\IG[width=0.33\textwidth]{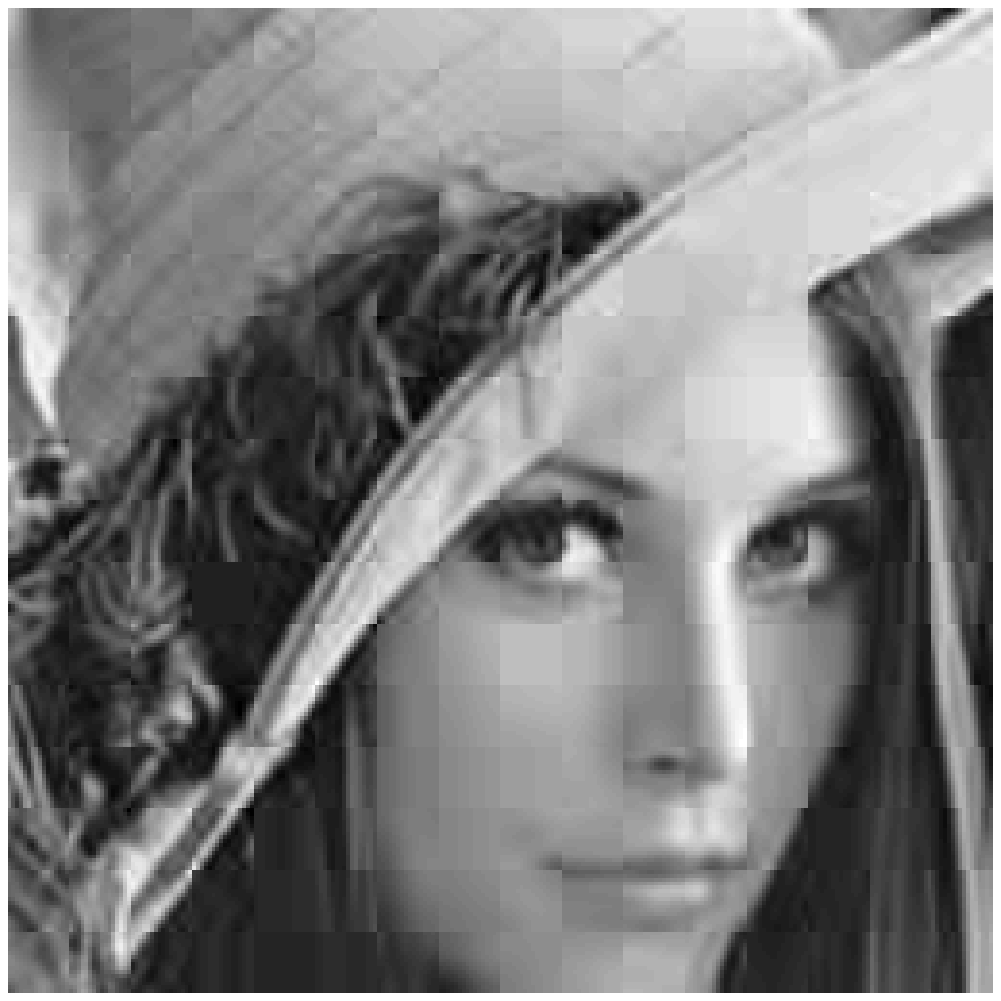}  &  \IG[width=0.33\textwidth]{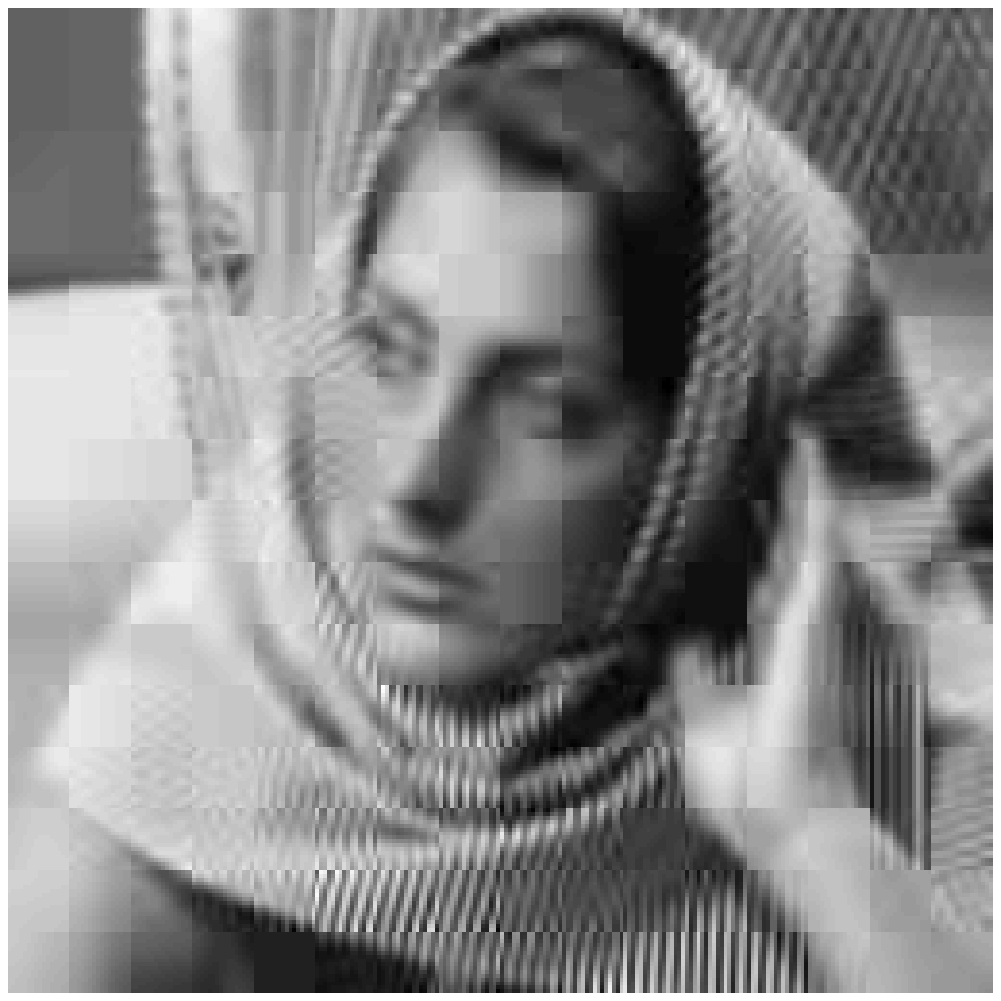}  \\
\IG[width=0.33\textwidth]{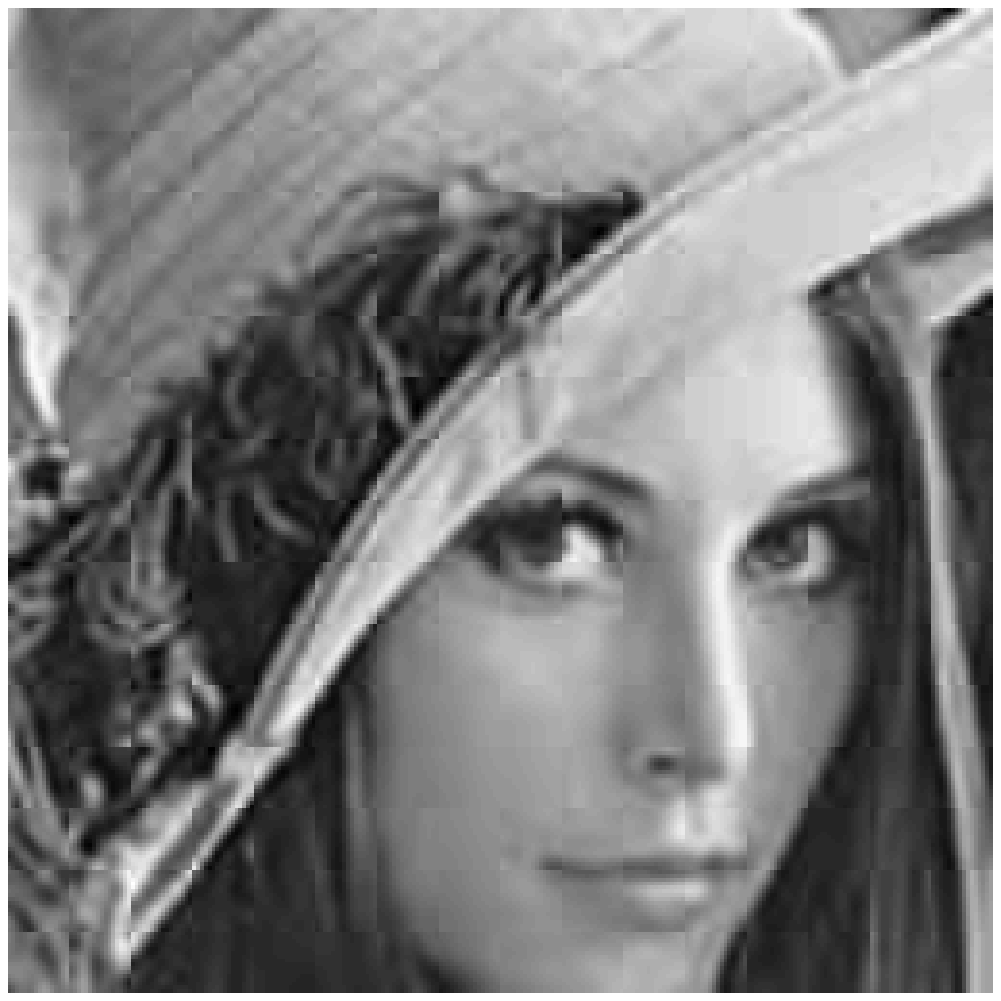}  &  \IG[width=0.33\textwidth]{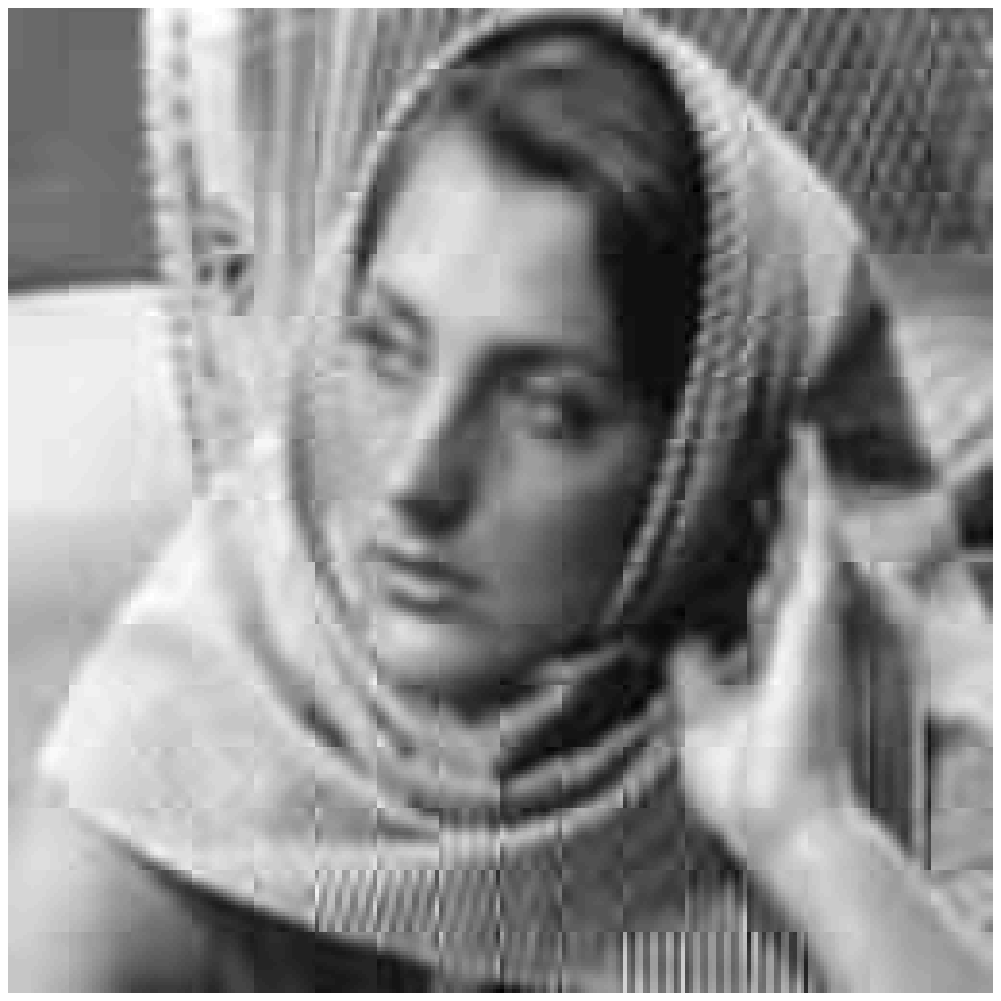}  \\
\IG[width=0.33\textwidth]{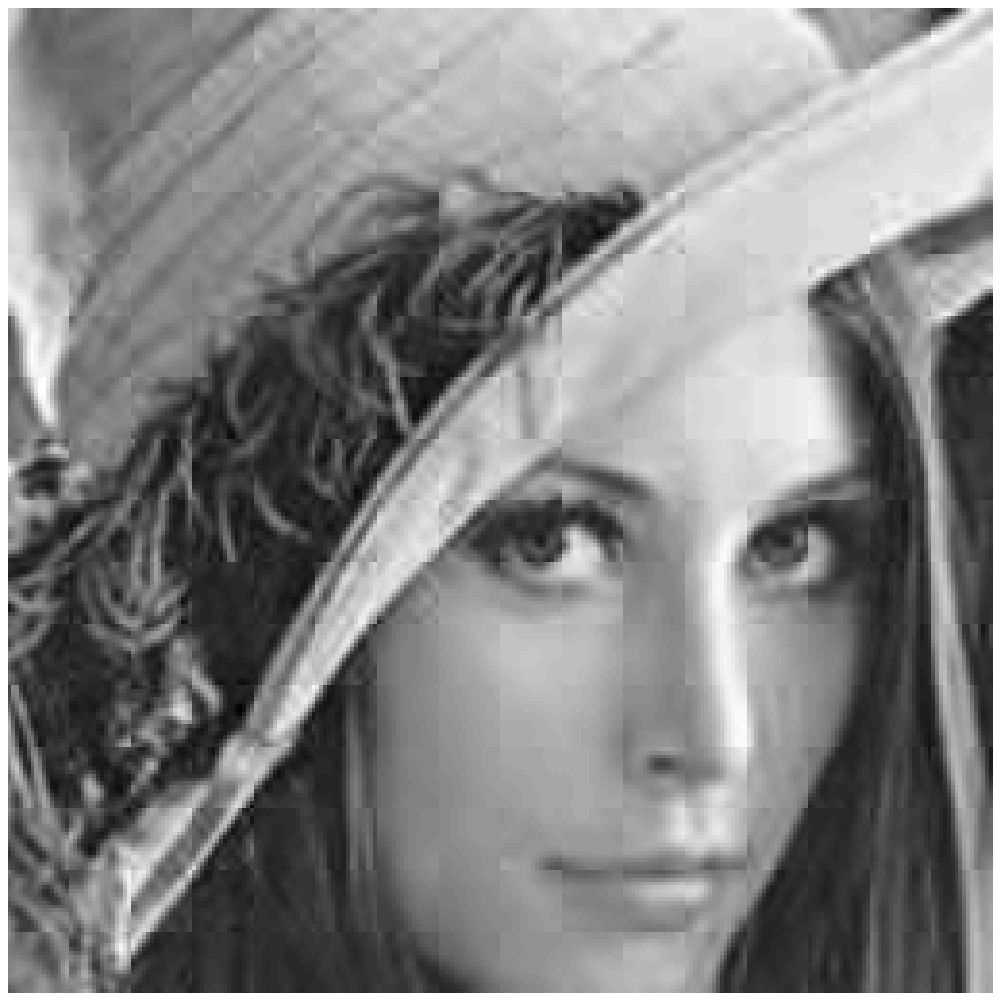} &  \IG[width=0.33\textwidth]{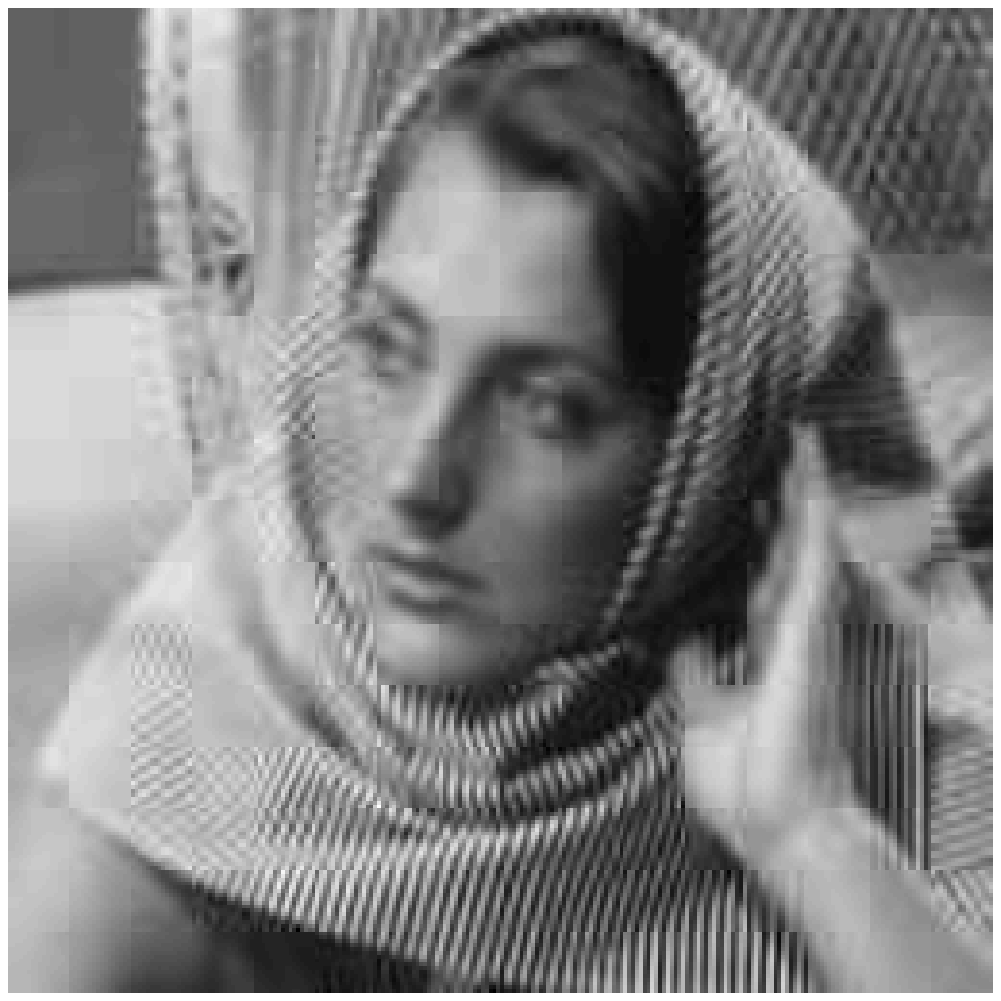}  \\
\end{tabular}
\end{center}
\vspace{-0.28in}
\caption{Examples of decoded Lena (left) and Barbara (right) images at 0.3
bits/pix. From top to bottom: JPEG, RKi-1, CSF-SVR, and NL-SVR.}
\label{images}
\end{figure}

Figure \ref{images} shows representative visual results of the considered
SVM strategies on standard images (Lena and Barbara) at the same
bit rate (0.3 bpp, 27:1 compression ratio or 2.4 KBytes in 256$\times$256 images).
The visual inspection confirms that the
\emph{numerical} gain in MPE and SSIM shown in Figure~\ref{rds} is
also \emph{perceptually significant}. Some conclusions can be
extracted from this figure. First, as previously reported
by \citet{Gomez05}, RKi-1 leads to poorer (blocky) results because
of the crude approximation of the CSF (as an ideal low-pass
filter) and the equal relevance applied to the low-frequency
DCT-coefficients. Second, despite the good performance yielded by
the CSF-SVR approach to avoid blocking effects, it is worth noting
that high frequency details are smoothed (e.g., see Barbara's
scarf). These effects are highly alleviated by introducing SVR in
the non-linear domain. See, for instance, Lena's eyes, her hat's
feathers or the better reproduction of the high frequency pattern
in Barbara's clothes.

\begin{figure}
\begin{center}
\begin{tabular}{cc}
{\it (a)} & {\it (b)}\\
\IG[width=0.33\textwidth]{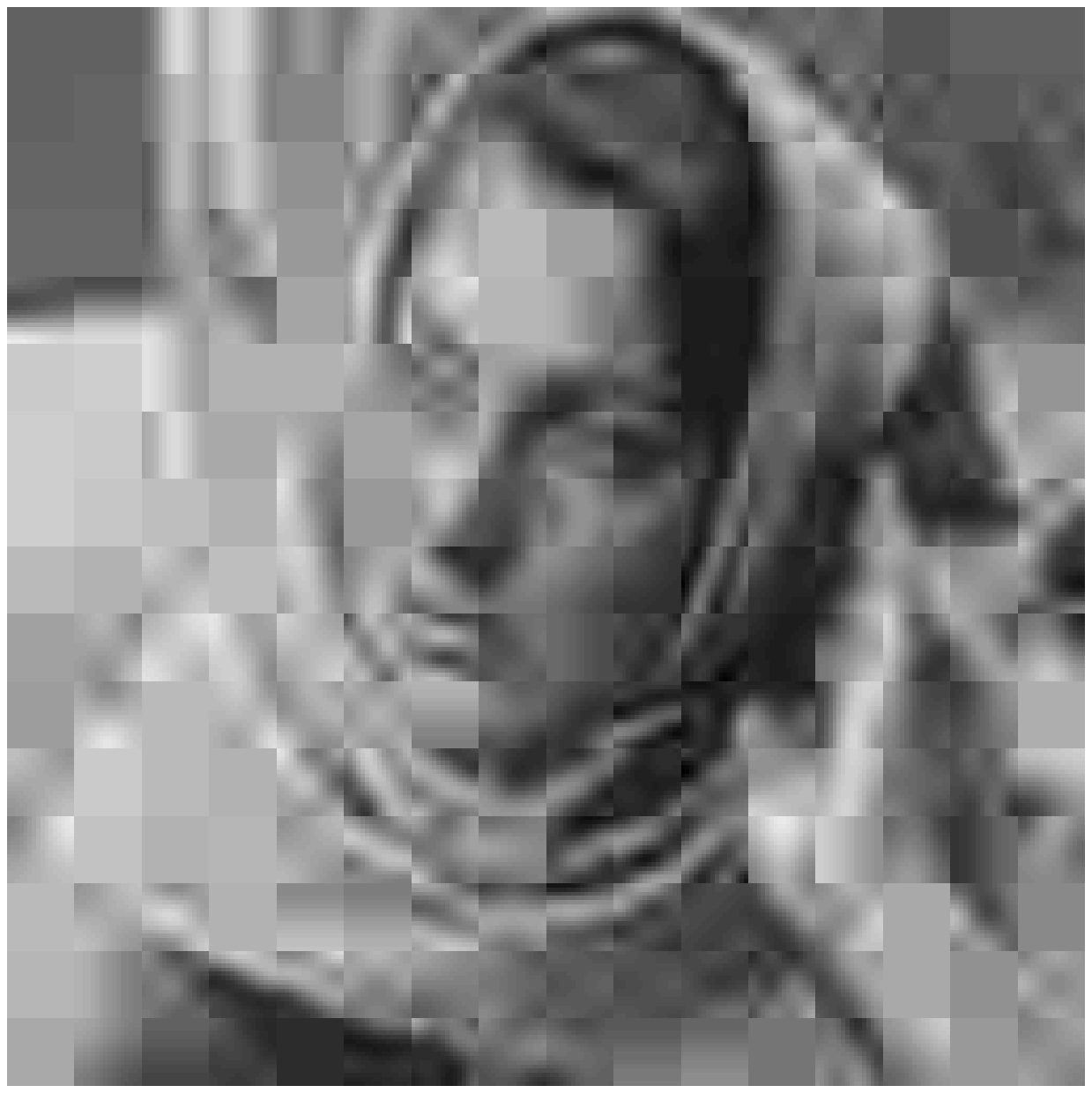} & \IG[width=0.33\textwidth]{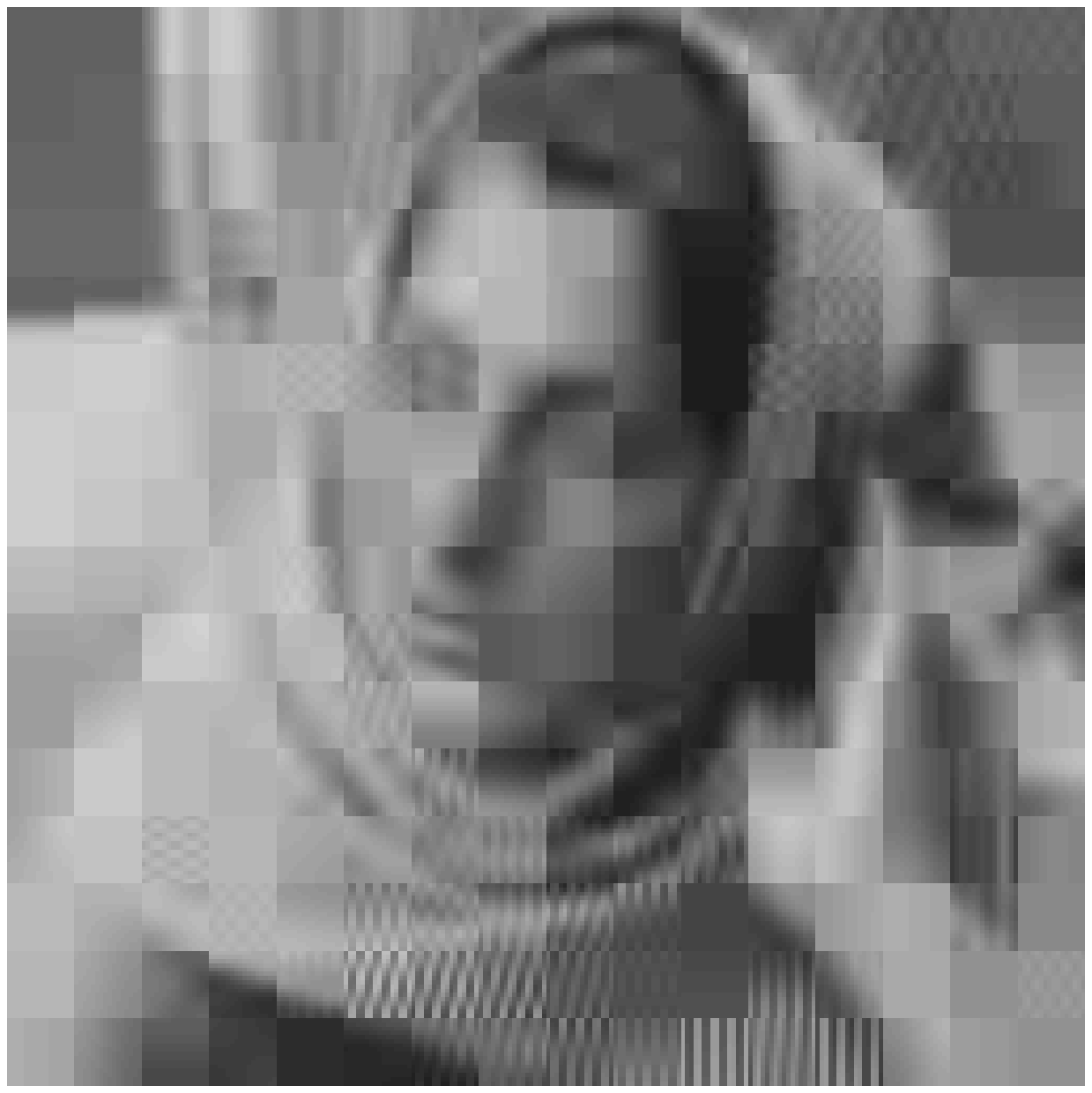} \\
{\it (c)} & {\it (d)}\\
\IG[width=0.33\textwidth]{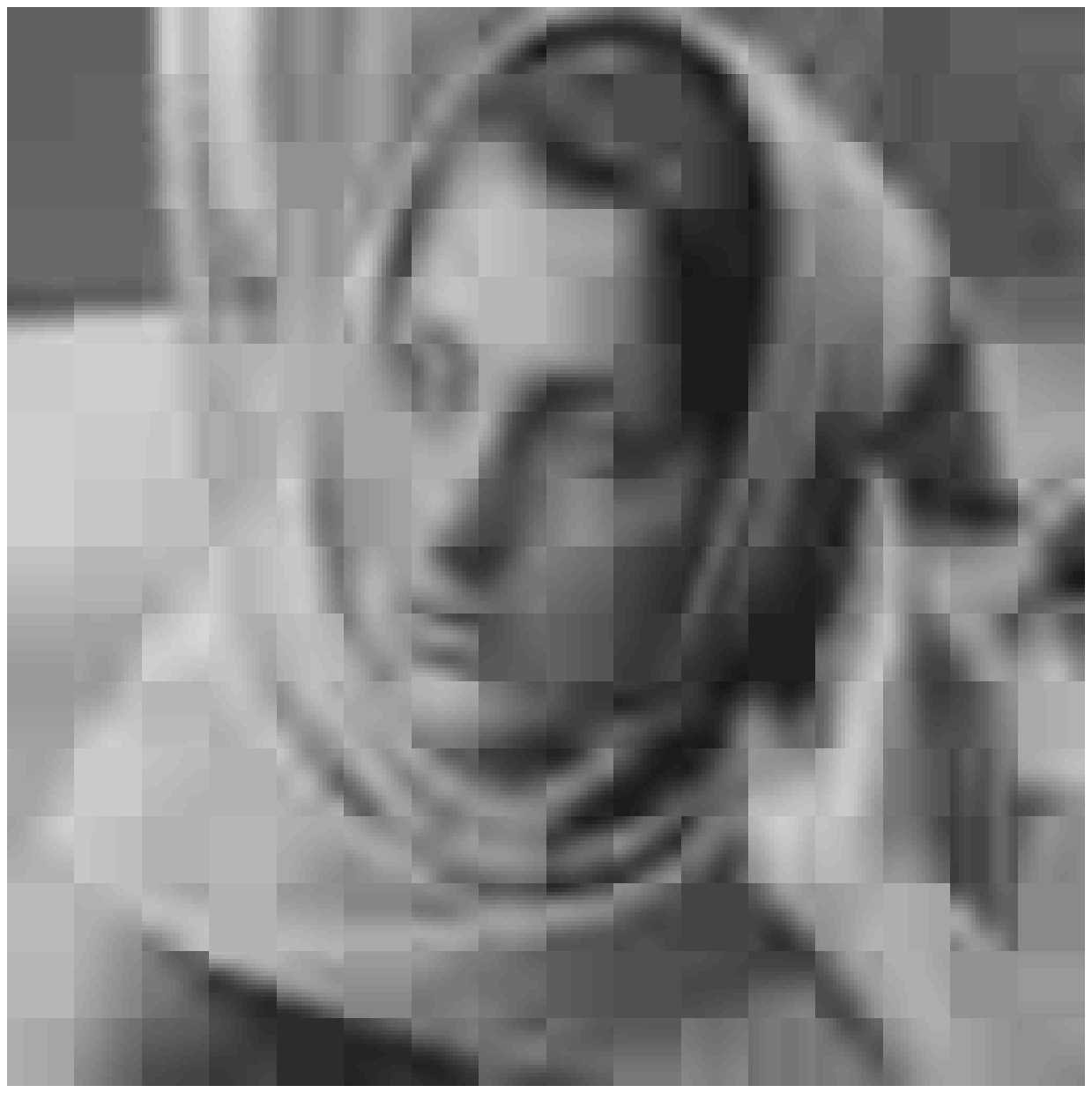} & \IG[width=0.33\textwidth]{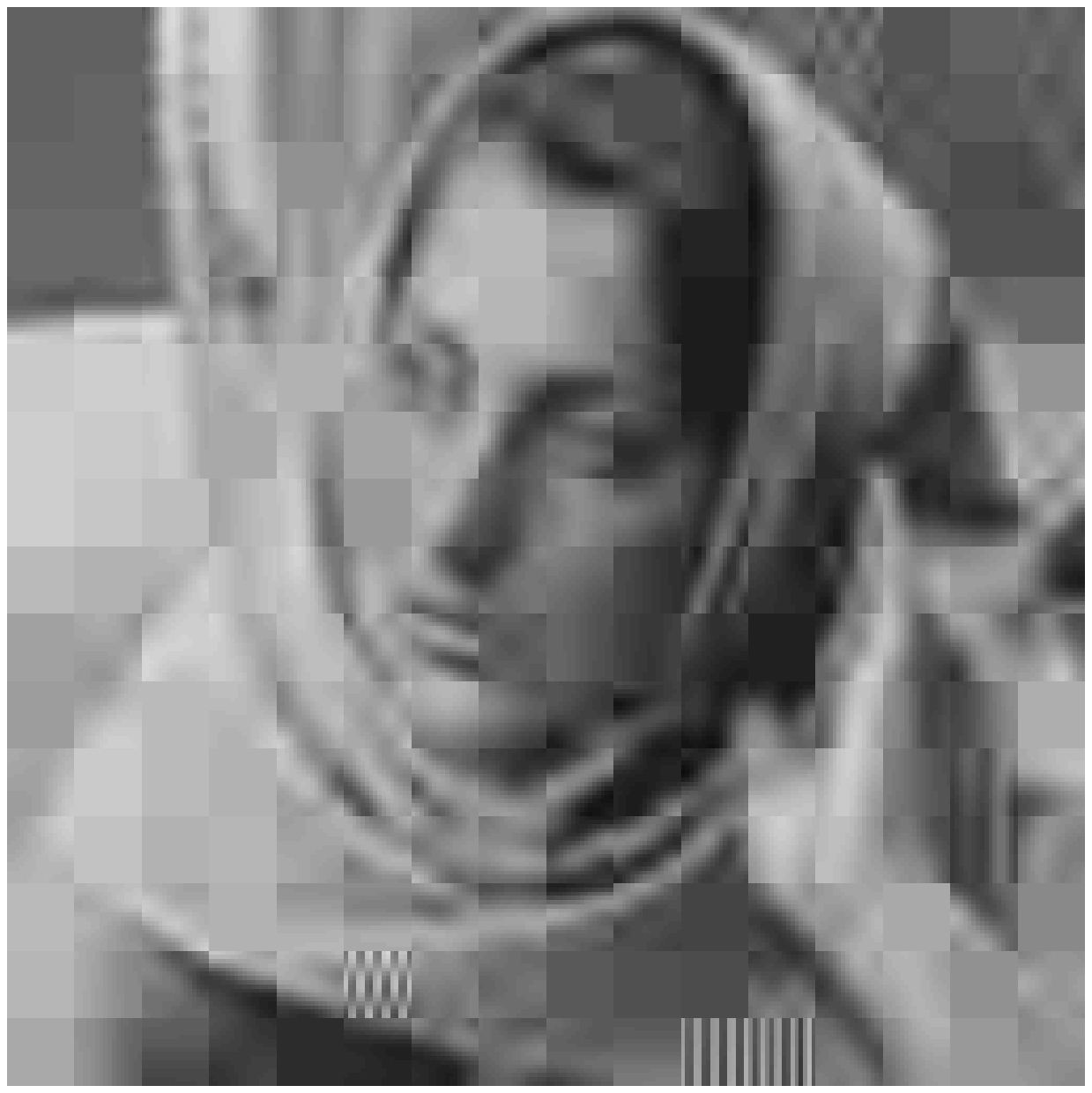}\\
\end{tabular}
\end{center}
\caption{Examples of decoded Barbara images at a high compression ratio of 0.05 bits/pix (160:1) for (a) JPEG, (b) RKi-1, (c) CSF-SVR, and (d) NL-SVR.}
\label{highrates}
\end{figure}

Figure \ref{highrates} shows the results obtained by all considered
methods at a very high compression ratio for the Barbara image
(0.05 bpp, 160:1 compression ratio or 0.4 KBytes in 256$\times$256 images).
This experiment is just intended to show the limits of
methods performance since it is out of the recommended rate ranges.
Even though this scenario is unrealistic, differences
among methods are still noticeable: the proposed NL-SVR
method reduces the blocky effects (note for instance that
the face is better reproduced). This is due to a better
distribution of support vectors in the perceptually independent domain.

\subsection{Support Vector Distribution}

\begin{figure}
\begin{center}
\begin{tabular}{ccc}
{\it (a)} & {\it (b)} & {\it (c)}\\
\IG[width=0.3\textwidth]{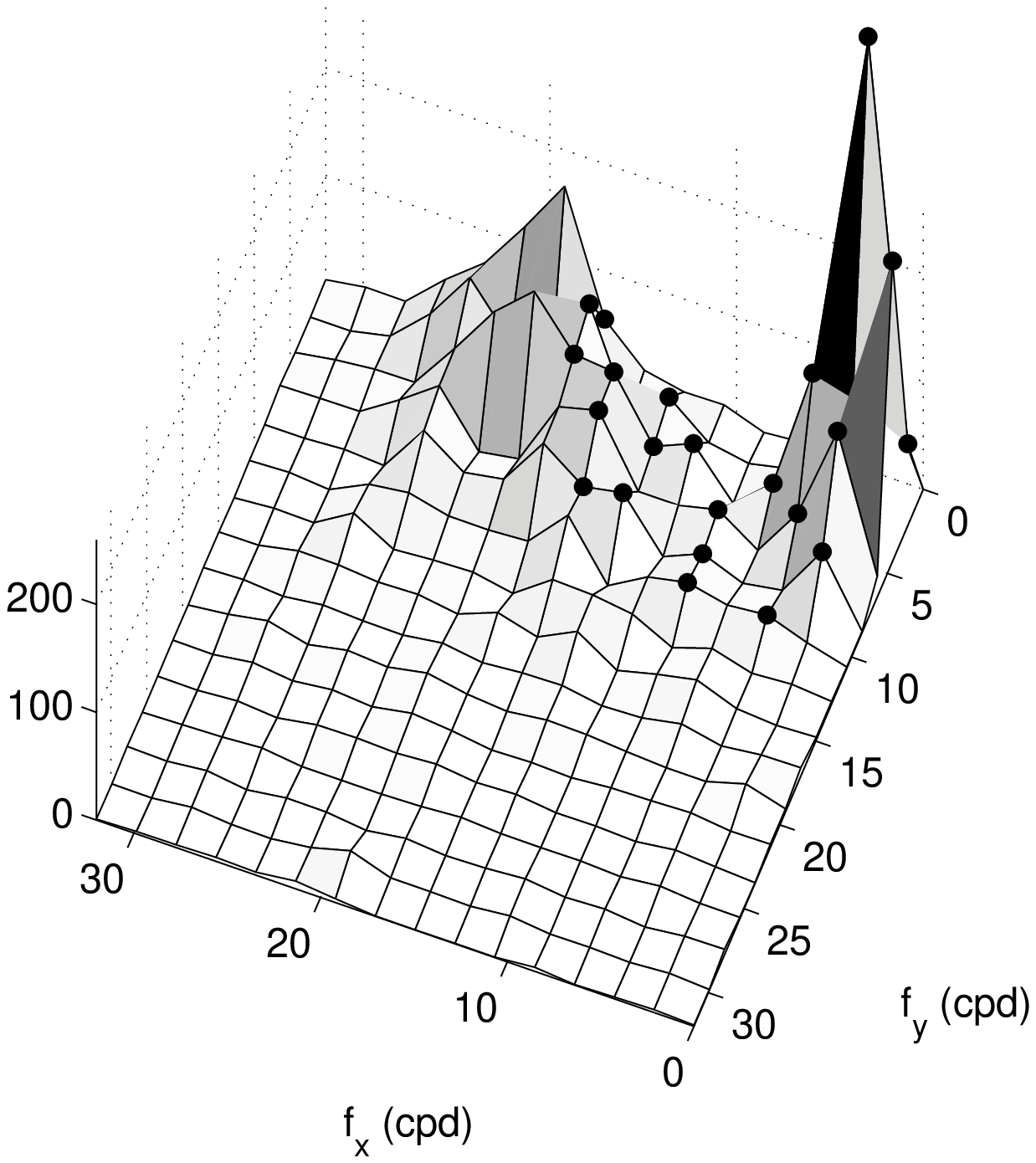}  &   \IG[width=0.3\textwidth]{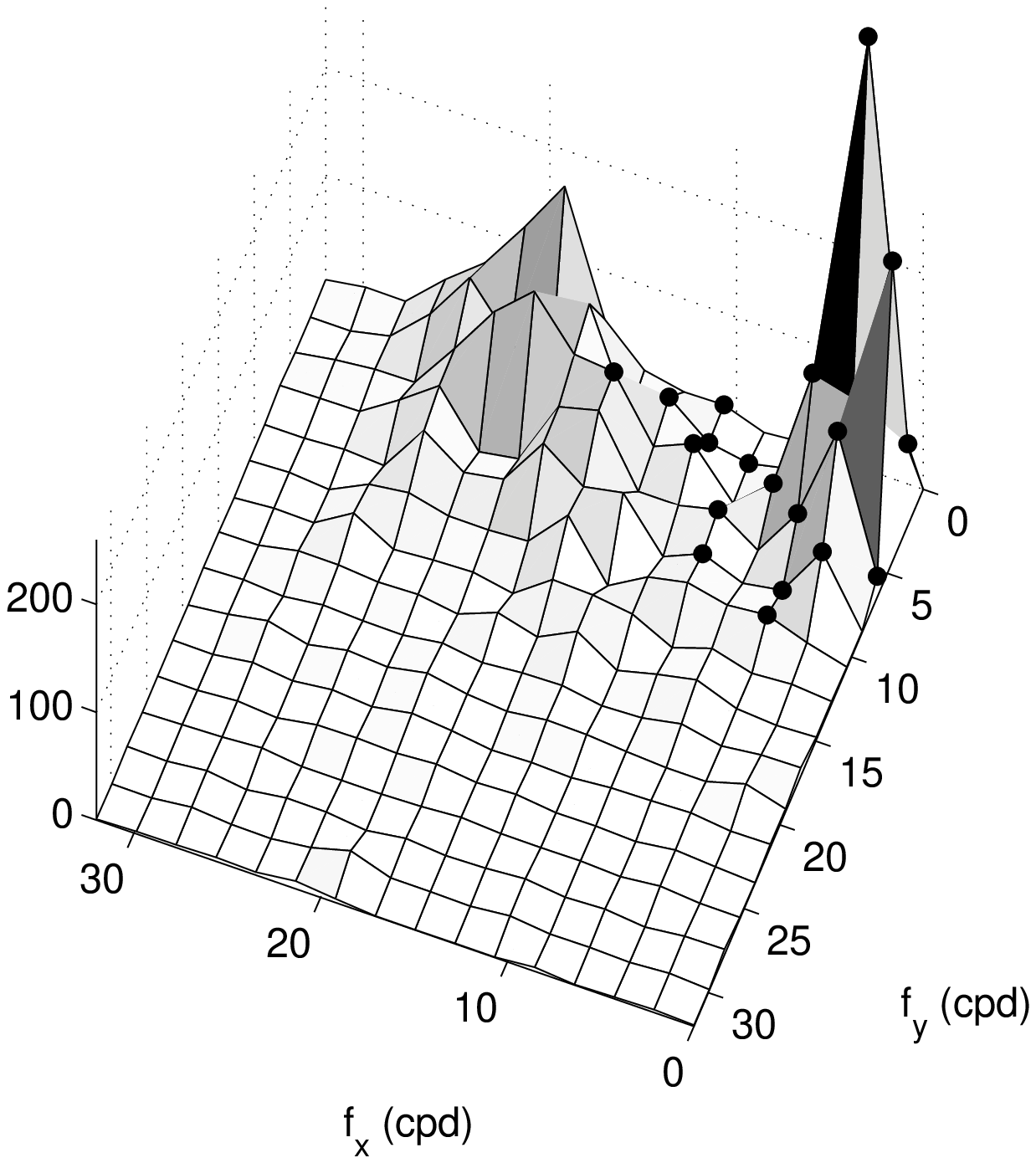}  &   \IG[width=0.3\textwidth]{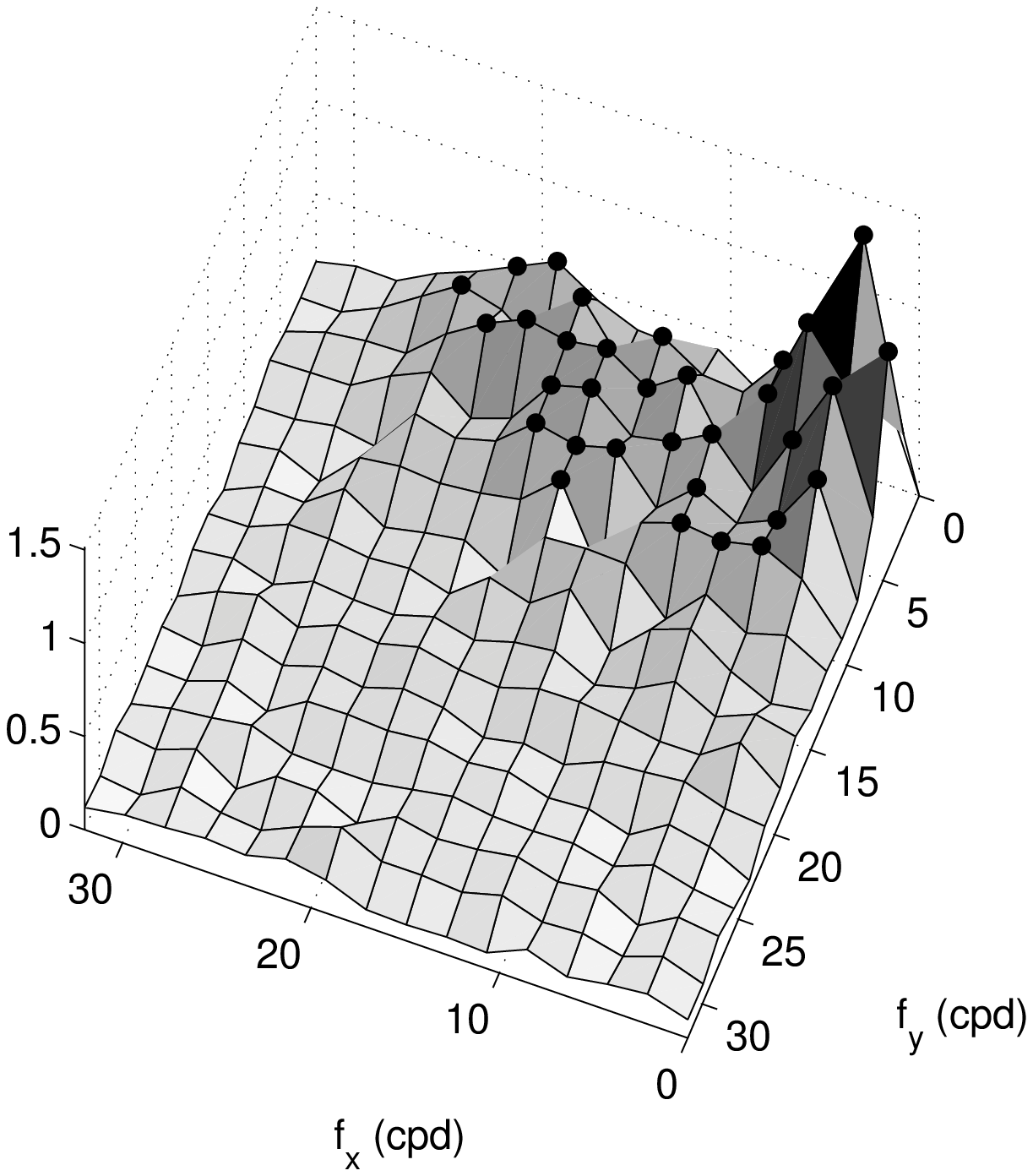}  \\
\IG[width=0.3\textwidth]{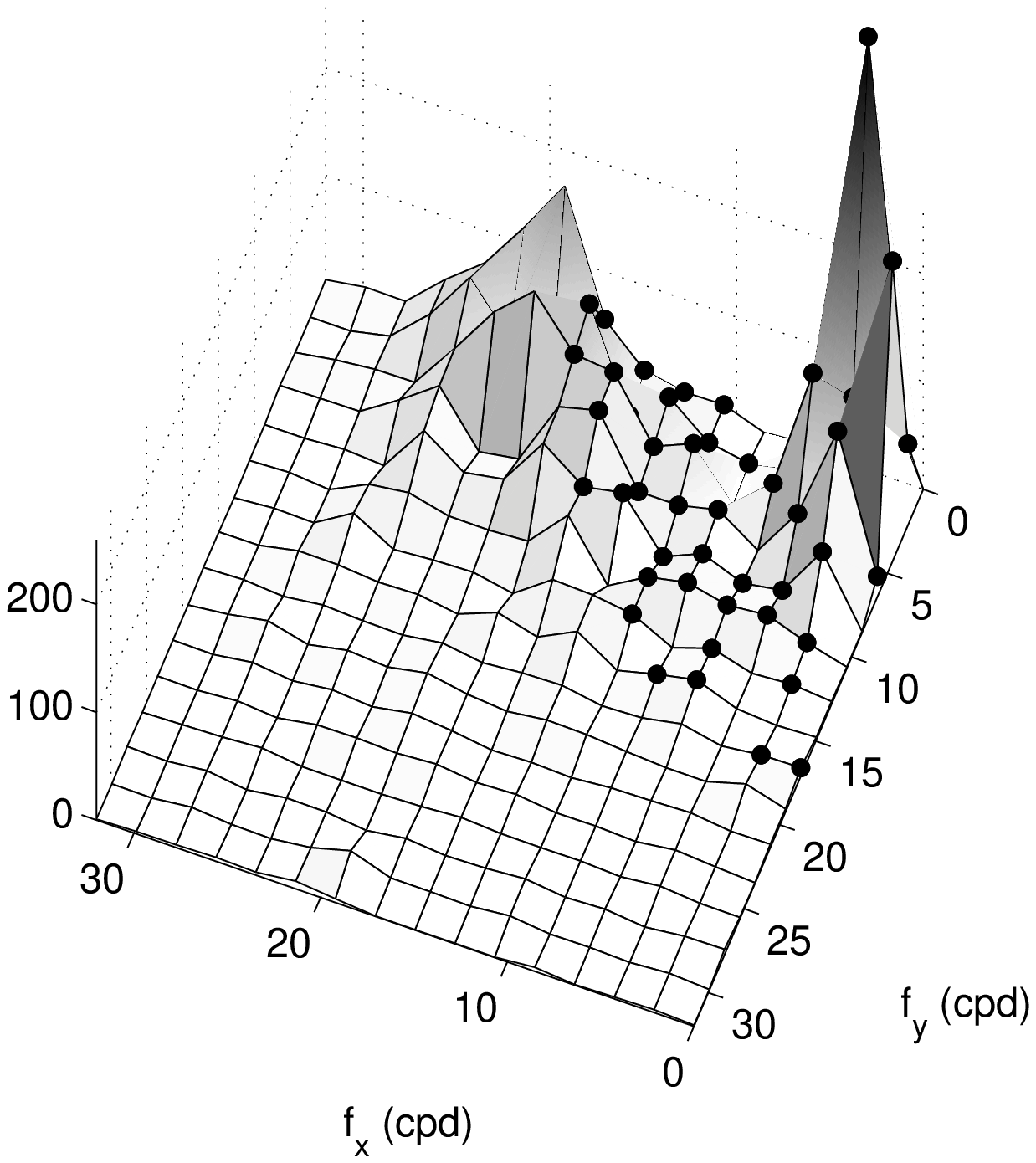}  &   \IG[width=0.3\textwidth]{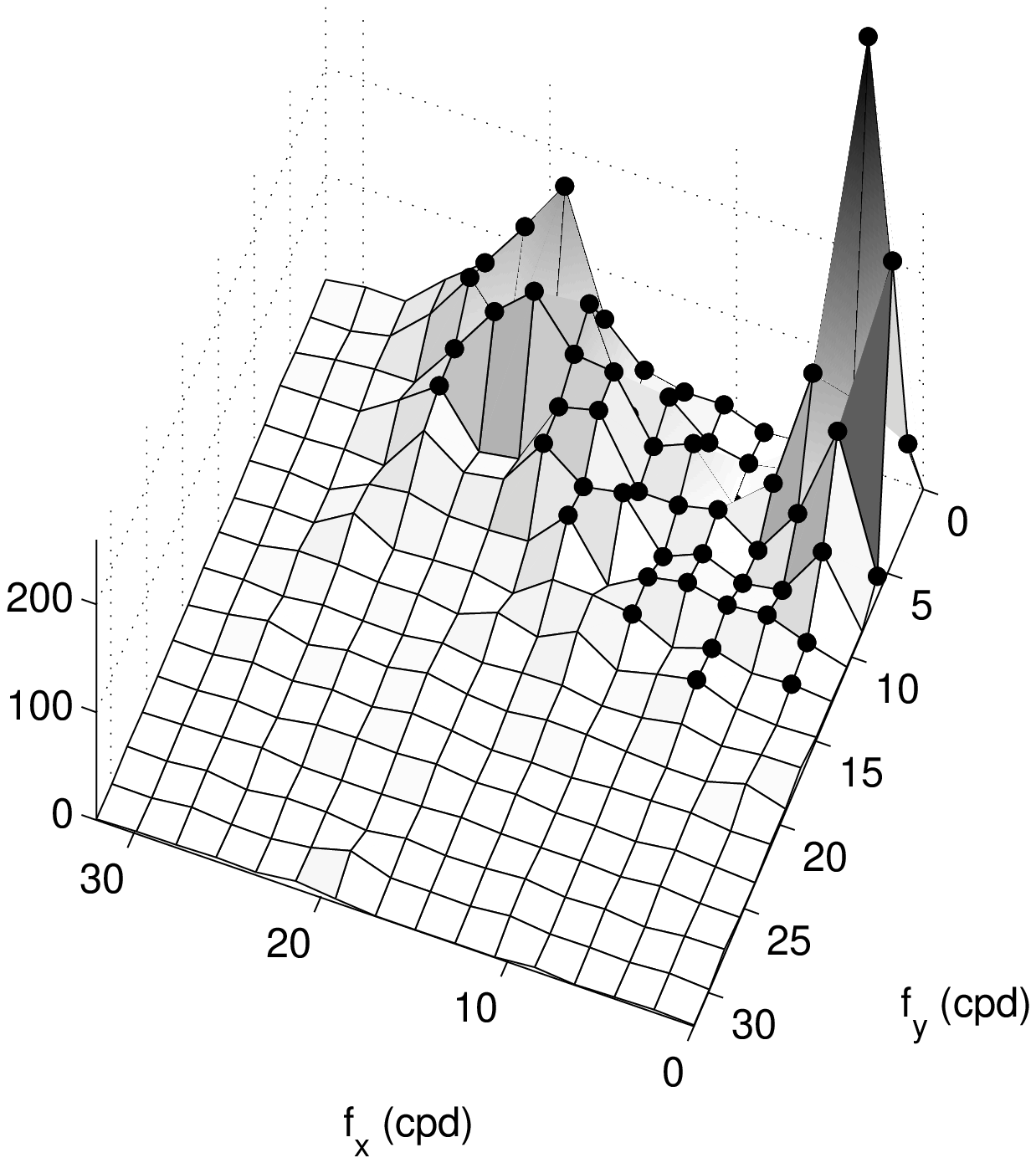}  &   \IG[width=0.3\textwidth]{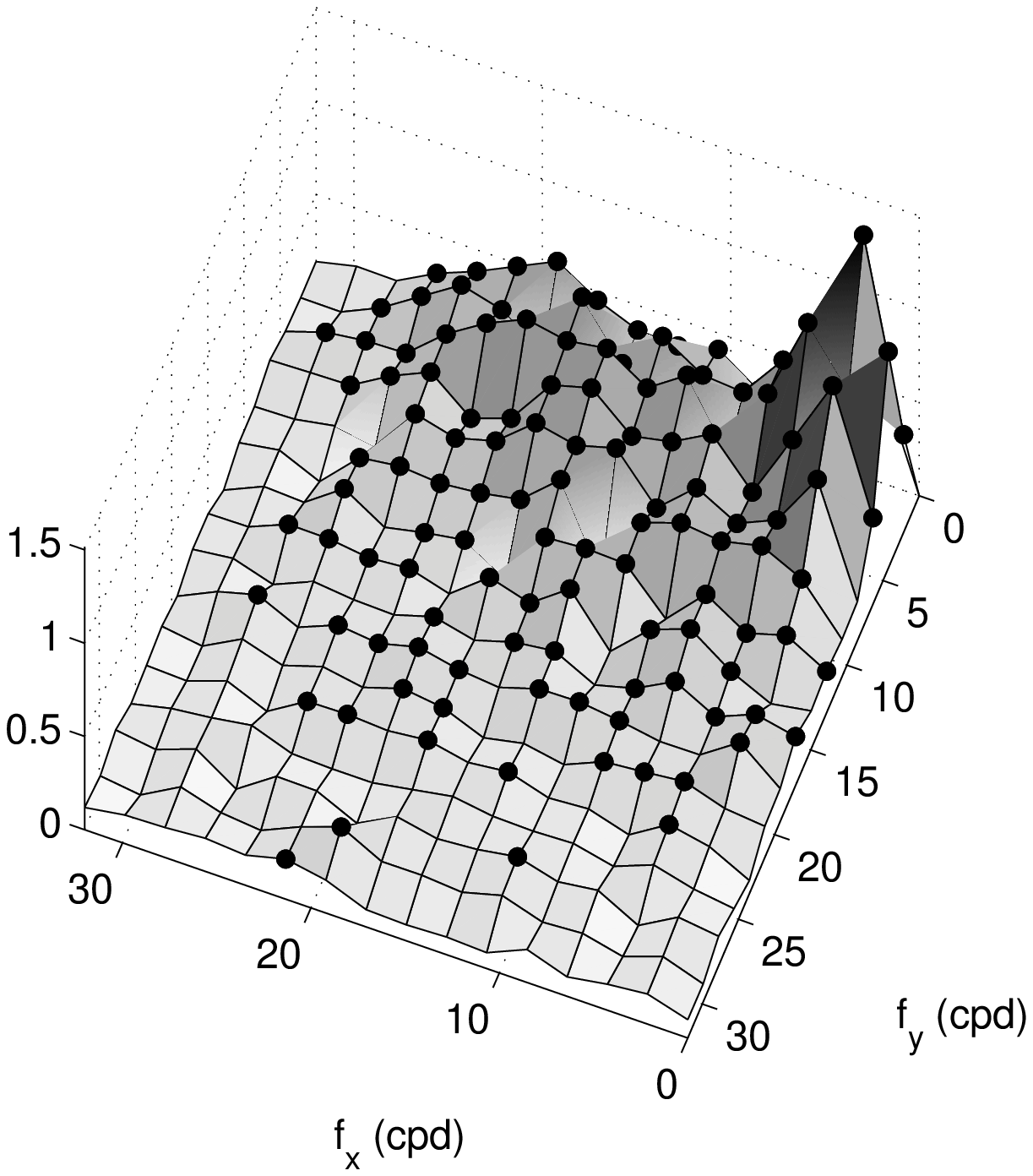}  \\
\end{tabular}
\end{center}
\caption{Signal in different domains and the selected support
vectors by the SVM models in a block of the Barbara image at 0.3
bits/pix (top row) and 0.5 bits/pix (bottom row). Different
domains are analyzed: (a) linear DCT using RKi-1, (b) linear DCT
with CSF-SVM, and (c) non-linear perceptual domain with standard
$\varepsilon$-SVM (NL-SVR).} \label{svs}
\end{figure}

The observed different perceptual image quality obtained with each
approach is a direct consequence of support vector distribution
in different domains. Figure \ref{svs} shows a representative example of the
distribution of the selected support vectors by the RKi-1 and the
CSF-SVR models working in the linear DCT domain, and the NL-SVM
working in the perceptually independent non-linear domain {\bf r}.
Specifically, a block of Barbara's scarf at different compression
ratios is used for illustration purposes.

The RKi-1 approach \citep{Robinson03} uses a constant
$\varepsilon$ but, in order to consider the low subjective
relevance of the high-frequency region, the corresponding
coefficients are neglected. As a result, this approach only
allocates support vectors in the low/medium frequency regions. The
CSF-SVR approach uses a variable $\varepsilon$ according to the
CSF and gives rise to a more natural concentration of support
vectors in the low/medium frequency region, which captures medium
to high frequency details at lower compression rates (0.5
bits/pix). Note that the number of support vectors is bigger than
in the RKi-1 approach, but it selects some necessary
high-frequency coefficients to keep the error below the selected
threshold. However, for bigger compression ratios (0.3 bits/pix),
it misrepresents some high frequency, yet relevant, features
(e.g., the peak from the stripes). The NL-SVM approach works in the
non-linear transform domain, in which a more uniform coverage of
the domain is done, accounting for richer (and perceptually
independent) coefficients to perform efficient sparse signal
reconstruction.

It is important to remark that, for a given method (or domain),
tightening $\varepsilon_f$ implies (1) considering more support
vectors, and (2) an increase in entropy (top and bottom rows in
Figure~\ref{svs}, 0.3 bpp to 0.5 bpp). However, note that the
relevant measure is the entropy and not the number of support
vectors: even though the number of selected support vectors in the
$\mathbf{r}$ domain is higher, their variance is lower, thus
giving rise to the same entropy after entropy coding.

\section{Conclusions}\label{conclusions}

In this paper, we have reported a condition on the suitable domain
for developing efficient SVM image coding schemes. The so-called
{\em diagonal Jacobian condition} states that SVM regression with
scalar-wise error restriction in a particular domain makes sense
only if the transform that maps this domain to an independent
coefficient representation is locally diagonal. We have
demonstrated that, in general, linear domains do not fulfill this
condition because non-trivial statistical and perceptual
inter-coefficient relations do exist in these domains.

This theoretical finding has been experimentally confirmed by
observing that improved compression results are obtained when SVM
is applied in a non-linear perceptual domain that starts from the
same linear domain used by previously reported SVM-based image
coding schemes. These results highlight the relevance of an
appropriate image representation choice before SVM learning.

Further work is tied to the use of SVM-based coding schemes in
statistically, rather than perceptually, independent non-linear
ICA domains. In order to do so, local PCA instead of local ICA may
be used in the local-to-global differential
approach \citep{Malo06b} to speed up the non-linear computation.

\acks{This work has been partly supported by the Spanish Ministry of
Education and Science under grant CICYT TEC2006-13845/TCM and by
the Generalitat Valenciana under grant GV-06/215.}

\bibliography{final}

\end{document}